\documentclass[10pt,journal,compsoc]{IEEEtran}

\ifCLASSOPTIONcompsoc
  \usepackage[nocompress]{cite}
\else
  \usepackage{cite}
\fi
\ifCLASSINFOpdf
\else
\fi
\usepackage{xcolor}
\usepackage{color}
\definecolor{mygray}{RGB}{193,203,215}
\usepackage{booktabs}
\usepackage{multirow}
\usepackage{colortbl}
\usepackage{graphicx}
\usepackage{caption}
\DeclareCaptionFont{blue}{\color{blue}}
\usepackage{booktabs}
\usepackage{array}
\usepackage{dblfloatfix}
\usepackage{balance}
\usepackage{enumitem}
\usepackage{url}
\usepackage{dingbat}
\newcolumntype{C}[1]{>{\centering\arraybackslash}m{#1}}

\usepackage[framemethod=tikz]{mdframed}

\usepackage{bigstrut}
\usepackage{multirow}
\usepackage{amsmath}
\usepackage{subfigure}

\makeatletter
\long\def\@IEEEtitleabstractindextextbox#1{\parbox{0.922\textwidth}{#1}}
\makeatother

\usepackage[rawfloats=true]{floatrow} 
\restylefloat{figure} 
\restylefloat{table} 

\usepackage{cite}
\usepackage{amsmath,amssymb,amsfonts}
\usepackage{algorithmic}
\usepackage{caption,graphicx}
\usepackage{textcomp}
\usepackage{xcolor}
\usepackage{mathrsfs}
\usepackage{pgfplots}

\usepackage{amssymb}
\usepackage{pifont}

\usepackage{subcaption}
\usepackage[table,xcdraw]{xcolor}
\usepackage{url}
\usepackage{textcomp}
\usepackage{xcolor}
\usepackage{mathrsfs}
\usepackage{booktabs} 
\usepackage{tikz}
\usepackage{dblfloatfix}
\usepackage{bigstrut}
\usepackage{multirow}
\usepackage{color}
\usepackage{fancyhdr}
\usepackage{listings} 
\usepackage{changepage}
\usepackage{lipsum}
\usepackage{balance}
\usepackage{enumitem}
\usepackage{comment}
\usepackage{breqn}
\usepackage{dblfloatfix}
\usepackage{booktabs} %
 \usepackage{makecell}

\begin{document}
%
\title{SpectraIrisPAD: Leveraging Vision Foundation Models for Spectrally Conditioned Multispectral Iris Presentation Attack Detection}
%
%
%
%

\author{
{\large Raghavendra Ramachandra$^1$  \ Sushma Venkatesh$^2$}\\
    {\normalsize
    $^1$ Norwegian University of Science and Technology (NTNU), Gj{\o}vik, Norway\\
    $^2$ MOBAI AS, Gj{\o}vik, Norway}
    }

%
%

\markboth{Journal of \LaTeX\ Class Files,~Vol.~14, No.~8, August~2015}%
{Shell \MakeLowercase{\textit{et al.}}: Bare Advanced Demo of IEEEtran.cls for IEEE Computer Society Journals}
%



\IEEEtitleabstractindextext{%
\begin{abstract}
Iris recognition is widely recognized as one of the most accurate biometric modalities. However, its growing deployment in real-world applications raises significant concerns regarding its vulnerability to Presentation Attacks (PAs). Effective Presentation Attack Detection (PAD) is therefore critical to ensure the integrity and security of iris-based biometric systems. While conventional iris recognition systems predominantly operate in the near-infrared (NIR) spectrum, multispectral imaging across multiple NIR bands provides complementary reflectance information that can enhance the generalizability of PAD methods.
In this work, we propose \textbf{SpectraIrisPAD}, a novel deep learning-based framework for robust multispectral iris PAD. The SpectraIrisPAD leverages a DINOv2 Vision Transformer (ViT) backbone equipped with learnable spectral positional encoding, token fusion, and contrastive learning to extract discriminative, band-specific features that effectively distinguish bona fide samples from various spoofing artifacts.
Furthermore, we introduce a new comprehensive  dataset Multispectral Iris PAD (\textbf{MSIrPAD}) with diverse PAIs, captured using a custom-designed multispectral iris sensor operating at five distinct NIR wavelengths (800\,nm, 830\,nm, 850\,nm, 870\,nm, and 980\,nm). The dataset includes 18,848 iris images encompassing eight diverse PAI categories, including five textured contact lenses, print attacks, and display-based attacks.
We conduct comprehensive experiments under unseen attack evaluation protocols to assess the generalization capability of the proposed method. SpectraIrisPAD consistently outperforms several state-of-the-art baselines across all performance metrics, demonstrating superior robustness and generalizability in detecting a wide range of presentation attacks.
\end{abstract}

}

\maketitle

\IEEEdisplaynontitleabstractindextext

%
\IEEEpeerreviewmaketitle

\section{Introduction}
\label{sec:introduction}
Iris recognition is considered one of the most accurate and reliable biometric technologies due to the inherent richness, stability, and distinctiveness of iris texture patterns. Unlike other biometric traits, the iris exhibits high inter-class variability and low intra-class variability, while remaining largely stable throughout an individual’s lifetime \cite{IrisSurevey2024deep}. These characteristics make it particularly well-suited for robust identity verification across a wide range of operational conditions. As a result, iris recognition has been widely adopted in both high-security domains such as border control and national identity programs, as well as in commercial applications including smartphone authentication and access control systems \cite{IrisSurevey2024deep, IrisSAmrtphone, IrisPAD_Survey}.
Most operational iris recognition systems acquire images in the Near Infrared (NIR) spectrum, where the absorption of light by melanin is significantly reduced. This spectral property enhances the visibility of intricate iris textures, particularly in individuals with dark irises, thereby supporting more accurate feature extraction. 
Given the widespread deployment of iris biometrics in critical infrastructures, it is imperative to understand their susceptibility to adversarial inputs, such as presentation attacks, and to develop effective countermeasures to ensure system reliability and long-term security.

With the growing deployment of iris biometrics, concerns about system vulnerability to covert and subversive attacks have intensified, particularly in the form of Presentation Attacks (PAs). NIR-based PAD techniques have been developed to counter such threats by identifying spoofing attempts using printed images, cosmetic contact lenses, and replayed videos \cite{IrisSurevey2024deep}. These methods typically exploit photometric inconsistencies, textural artifacts, and dynamic cues such as pupil reactivity \cite{IrisSurevey2024deep, IrisSAmrtphone, IrisPAD_Survey}. Despite these efforts, existing PAD approaches often suffer from limited generalization across sensors, environmental conditions, and unseen attack types. Results from the LivDet-Iris competitions~\cite{tinsley2023iris} have underscored these limitations, revealing a significant performance drop when PAD systems are evaluated on previously unseen presentation attack instruments or datasets. These findings highlight the need for more robust and spectrally adaptive PAD frameworks capable of defending against diverse and evolving attack vectors.

Conventional PAD approaches often rely on single-spectrum Near-Infrared (NIR) imaging, which may fail to capture sufficient discriminative cues to generalize across unseen attack types~\cite{tinsley2023iris}. Recent studies have shown that the spectral reflectance properties of bona fide irises and attack instruments vary significantly across wavelengths~\cite{raghavendraMSIrisSensor}. This motivates the integration of multispectral imaging into PAD systems, as it enables the capture of complementary spectral signatures that are otherwise not observable in standard NIR images. As demonstrated in prior work~\cite{raghavendraMSIrisSensor,iris_MS_sensor, ramachandra2024multi, MSIrisCommercial}, multispectral sensors operating across a range of NIR bands (e.g., 800\,nm to 980\,nm) can enhance both verification and PAD performance. These sensors facilitate the detection of subtle spectral differences between real and fake irises by exploiting wavelength-specific light absorption and scattering characteristics. Furthermore, the use of multispectral data offers better robustness against environmental variations and increases generalization capability when encountering novel or unseen PAIs. Therefore, leveraging multispectral imaging is a promising direction toward building more resilient and secure iris recognition systems capable of addressing evolving threats.

\subsection{Related Work on NIR Iris PAD}
Presentation Attack Detection (PAD) has been extensively studied in the context of iris biometrics, particularly for systems operating in the Near-Infrared (NIR) spectrum, due to their widespread deployment in commercial and security-critical applications.
Existing PAD methods for detecting the PAs in iris recognition systems can be broadly classified into the following five categories \cite{IrisSurevey2024deep, IrisPAD_Survey}. (1) \textit{Texture-based methods}, which analyze spatial texture patterns to distinguish between authentic irises and textured lenses~\cite{IITDDB, Daugirisantispoof, NDDB, Gragnaniello2015, Raghavendra:2014:ESI:2683483.2683507, zhang2010contact, 7264974, kohli2013revisiting, komulainen2014generalized, yadav2018fusion, GLCM_IrisSpoof}; (2) \textit{Image quality-based approaches}, which exploit degradation or inconsistencies in image quality metrics as cues for detecting presentation attacks~\cite{TIPIrisFace}; (3) \textit{Deep learning-based techniques}, which leverage convolutional neural networks or transfer learning to learn discriminative features directly from iris images~\cite{DeepImage, wei2008counterfeit, raghavendra2017contlensnet, DLTrnasferLearning_CLDetection}; (4) \textit{3D or meta-information-based strategies}, which utilize depth cues or structural information to differentiate real irises from spoof artifacts~\cite{3DFAkeIris, OCT_IrisPAD}. 

Deep learning has significantly advanced iris Presentation Attack Detection (PAD) by outperforming traditional handcrafted approaches. Prevailing strategies include: (a) utilizing pre-trained convolutional neural networks (CNNs) for feature extraction \cite{fang2021cross}, (b) implementing end-to-end learning architectures that directly optimize PAD classification objectives~\cite{jaswal2024learning},~\cite{agarwal2022enhanced},~\cite{agarwal2022generalizedIEEE},~\cite{sharma2025cascading},~\cite{choudhary2023identifying}, \cite{raghavendra2017contlensnet}and (c) employing generative adversarial networks (GANs) to generate synthetic iris samples, which augment limited training data and improve model generalization by training discriminators to recognize subtle attack patterns \cite{parametric_DNetPAD}.  Recent studies have explored large vision backbones such as DINOv2 and CLIP for iris PAD \cite{sony2025benchmarking}, demonstrating the potential of foundation models for biometric security. However, these works have been restricted to single-spectrum NIR imagery and lack explicit mechanisms to encode spectral semantics or to fuse features across different bands. In addition, their analyses are confined to zero-shot-æ¨´å´å+¨ evaluation, without assessing fine-tuned performance or cross-artefact generalisation. Despite the reasonable detection performance, existing NIR-based PAD approaches face several limitations: (a) limited generalization across different artifact types, as evidenced by poor cross-artefact performance; (b) reduced reliability when encountering novel or unseen Presentation Attack Instruments (PAI); (c) dependence on large-scale annotated datasets to effectively train end-to-end deep models; and (d) potential overfitting to dataset-specific cues, which undermines robustness in real-world deployment scenarios.

\subsection{Related Work on Multispectral Iris PAD}
Several studies have explored the utility of multispectral imaging for iris presentation attack detection. An early approach utilized Purkinje image reflections generated under two spectral bands, 760\,nm and 880\,nm, to form higher-order reflections for detecting artifacts such as 3D artificial eyes and patterned contact lenses~\cite{purkinje_Image}. While this method showed potential on a limited dataset, its effectiveness is constrained by the difficulty in reliably capturing higher-order reflections, particularly when contact lenses are worn~\cite{3DcontlensnetAdam}. Another work employed gradient-based feature fusion across four spectral bands (780\,nm, 810\,nm, 850\,nm, and 900\,nm), demonstrating promising results on print attack detection using a fused spectral representation~\cite{MS_Iris_PAD}. Similarly, reflectance-based features extracted from the iris-sclera boundary at 750\,nm and 850\,nm were shown to be useful for detecting both printed and textured contact lenses~\cite{MS_Iris_PAD_contactLens}; however, this method is sensitive to accurate segmentation of the sclera region, and performance may degrade if localization errors occur.

A custom-built multispectral iris sensor was also developed to detect prosthetic eyeballs composed of various materials~\cite{raghavendraMSIrisSensor}. Although this method achieved success in isolating attacks based on prosthetic eyes, its real-world applicability remains limited, as such artifacts are rarely encountered and generally fail to resemble the detailed texture of genuine irises, especially for individuals with darker irises. More recently, a deep learning-based PAD framework was proposed that captures iris images across five spectral bands from 800\,nm to 980\,nm and employs CNN-derived features for detecting two different patterned contact lens attacks~\cite{ramachandra2024multi}. This approach demonstrates superior performance compared to conventional NIR-only systems by leveraging spectral diversity and wavelength-specific discriminative cues. However, its generalization to diverse attack types and unseen artifacts remains an open challenge, highlighting the need for more comprehensive evaluation across heterogeneous presentation attack instruments.

\subsection{Our Contributions}

To address the limitations of existing multispectral iris PAD techniques, particularly their limited generalizability and robustness to diverse presentation attack instruments (PAIs), we present a comprehensive framework that advances both algorithm design and dataset availability. Existing studies often focus on narrow attack categories or evaluate performance using limited protocols, thus failing to establish generalization to unseen PAIs. Motivated by these challenges, this work introduces a new dataset and a novel deep learning-based PAD framework tailored for robust generalization across diverse attack types. First, we significantly extend our previous dataset~\cite{ramachandra2024multi} to include a broader spectrum of PAIs and a larger number of samples. Building on this, we propose a new architecture, termed \textbf{SpectraIrisPAD}, which leverages a DINOv2 backbone enhanced with spectral position encoding and token fusion mechanisms. These components are specifically designed to capture band-specific discriminative features that are crucial for detecting artefacts across varying PAIs.
Furthermore, we introduce a new multispectral iris PAD dataset, named \textbf{MSIrPAD}, comprising 18,848 iris images (3,535 bona fide and 15,313 attack samples), collected across five spectral bands (800\,nm, 830\,nm, 850\,nm, 870\,nm, and 980\,nm) using a custom multispectral iris sensor~\cite{raghavendraMSIrisSensor}. The dataset includes eight distinct PAIs, covering five textured contact lens variants, a Kindle display attack, and a print attack. To the best of our knowledge, MSIrPAD is the first multispectral iris dataset that offers this level of diversity and scale in attack instrumentation.

The main contributions of this work are summarized as follows:

\begin{enumerate}
    \item We propose \textbf{SpectraIrisPAD}, a novel multispectral iris PAD framework based on the DINOv2 backbone, integrated with learnable spectral position encoding and token fusion. This architecture is tailored to effectively capture spectral-specific features and enhance robustness against unseen presentation attacks.
    
    \item We introduce \textbf{MSIrPAD}, a new multispectral iris PAD dataset with 18,848 samples across five spectral bands. It includes a comprehensive set of eight presentation attacks and is the first of its kind in terms of diversity and sensor design.
    
    \item We conduct extensive experiments to benchmark the performance of SpectraIrisPAD against existing state-of-the-art methods. Evaluation protocols are designed to evaluate generalization to unseen attacks. The comparative analysis is provided with strong baselines, including \textit{DeFu}~\cite{ramachandra2024multi}, \textit{DNetPAD}~\cite{DNetPAD}, \textit{ADV-GEN}~\cite{parametric_DNetPAD}, \textit{CLIP}~\cite{sony2025benchmarking}, and \textit{ViTPAD}~\cite{sharma2025cascading}.
    
    \item The implementation code will be made publicly available to support reproducibility. The dataset, due to ethical and institutional constraints, can be shared upon reasonable request, subject to approval by the institutional ethics board.
\end{enumerate}
The remainder of this paper is structured as follows. Section~\ref{sec:Pro} introduces the proposed \textit{SpectraIrisPAD} framework for robust multispectral iris presentation attack detection. Section~\ref{sec:Db} describes the \textit{MSIrPAD} dataset, including the acquisition protocol,  statistical performance of the collected samples and ablation study. Section~\ref{sec:Exp} presents the experimental evaluation and benchmarking results in comparison with existing state-of-the-art methods and \label{sec:theory} discuss the theoretical justifications of the proposed SpectraIrisPAD.  Section \ref{sec:lim} discuss the limitation and future work. 
Finally, Section~\ref{sec:Conc} concludes the paper and outlines future directions.

\section{SpectraIrisPAD}
\label{sec:Pro}
\begin{figure*}[htp]
\begin{center}
\includegraphics[width=1.0\linewidth]{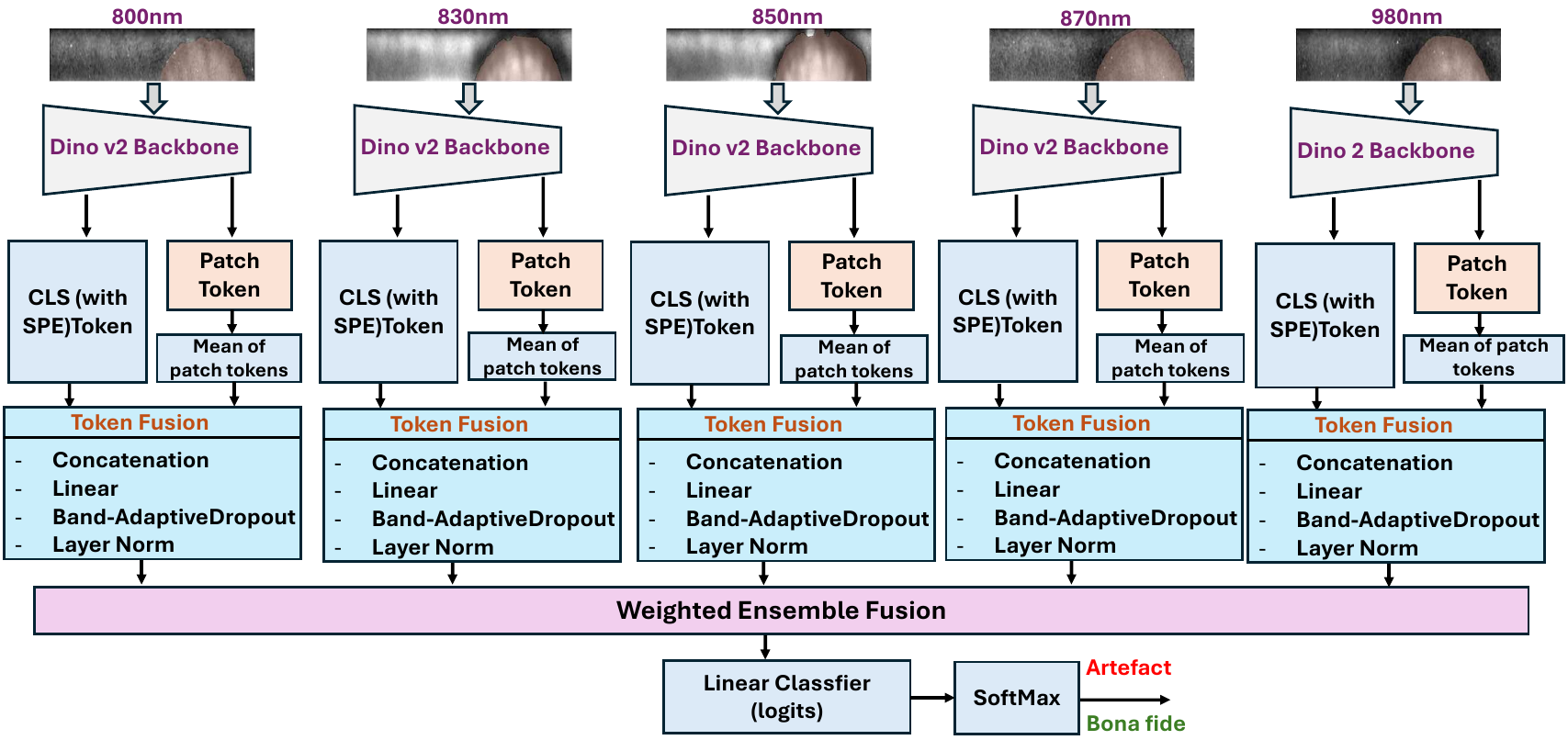}
\end{center}
      \caption{Overview of the proposed \textsc{SpectraIrisPAD}. Each band $b\in\{800,830,850,870,980\}$ is processed with DINOv2 and
\emph{Spectral Positional Encoding} by injecting a learnable embedding $E_b$ into the CLS token to obtain
$\mathrm{CLS}^{\mathrm{SPE}}_b$. The CLS and mean patch tokens are fused, followed by band-adaptive dropout (or BandDropout) with
probability $p_b$ and a linear classifier to produce the per-band posterior $P_b$. Bands are combined only at the
probability level using development accuracy weights $w_b$ to yield the fused prediction $P_{\mathrm{ens}}(x)$. Bands are fused with a mask-aware (missing-band robust) probability-level
ensemble, using development-set weights $w_b$. All
trainable modules are \emph{band-specific} with no cross-band weight sharing.} 
\vspace{4pt}
  {\footnotesize
  \textbf{Legend:} shared $\theta_{\text{shared}}$; band-specific heads $\phi_b$ (no sharing);
  SPE on CLS: “CLS $+\mathbf{e}_b$”; quality-aware gate: $g_b=f_q(x_b)$;
  fusion: $\hat{p}(y)=\sum_b \alpha_b\,p_b(y\!\mid\!x_b)$ (robust to missing bands).
  }
\label{fig:Prop}
\end{figure*}
Figure \ref{fig:Prop} illustrates the proposed SpectraIrisPAD a novel deep learning framework designed specifically for multispectral iris PAD, leveraging the unique spectral properties of iris images captured across multiple wavelength bands (800nm, 830nm, 850nm, 870nm, 980nm). The proposed \textbf{SpectraIrisPAD} framework begins by capturing multispectral iris images across five wavelength bands (800\,nm, 830\,nm, 850\,nm, 870\,nm, and 980\,nm). For each spectral image \( I_k \), where \( k \in \{1,2,3,4,5\} \), the iris region is segmented and normalized using the open-source OSIRIS v4.1 software~\cite{OTHMAN2016124}. This process results in standardized iris images \( N_s \), each of size \( 512 \times 64 \), which serve as the input to our deep learning pipeline. As illustrated in Figure~\ref{fig:Prop}, \textbf{SpectraIrisPAD} is specifically designed for multispectral iris PAD, leveraging the distinct spectral properties captured across these bands.
The SpectraIrisPAD leverages a DINOv2 backbone (ViT-B/14) \cite{oquab2023dinov2}, spectral positional encoding (SPE), token fusion, contrastive learning, band (or spectral band) adaptive dropout, class-balanced loss, feature normalization, and ensemble weighting for reliable multi-spectral iris PAD. 
In contrast to existing multispectral iris PAD methods that rely on CNN-based fusion or handcrafted features, our approach integrates spectral positional encoding, adaptive token fusion of ViT outputs, class-balanced contrastive training, and dynamic regularization. We further introduce a lightweight ensemble strategy leveraging per-band validation accuracy, enabling robust cross-spectrum generalization. To the best of our knowledge, this is the first work that systematically combines all these innovations within a unified DINOv2-based iris PAD framework. In the following, we provide a detailed description of each computational block in the proposed SpectraIrisPAD framework.

\subsection{Input: Multispectral Iris Images}
Given five spectral iris images for wavelength bands \(\lambda \in \{800, 830, 850, 870, 980\}\) nm, each image for band \(k \in \{1, 2, 3, 4, 5\}\) is denoted as:
\begin{equation}
I_k \in \mathbb{R}^{H \times W}
\end{equation}
where \(H = W = 224\) (after resizing), and \(I_k\) is a grayscale image normalized to \([0, 1]\). The images are converted to 3-channel inputs by repeating the grayscale channel as follows:
\begin{equation}
I_k^{\text{RGB}} = \text{repeat}(I_k, 3) \in \mathbb{R}^{3 \times H \times W}
\end{equation}
The images are normalized using band-specific statistics (mean \(\mu_k\), standard deviation \(\sigma_k\)) computed from the training set:
\begin{equation}
I_k^{\text{norm}} = \frac{I_k^{\text{RGB}} - \mu_k}{\sigma_k} \in \mathbb{R}^{3 \times H \times W}
\end{equation}

\subsection{DINOv2 Backbone}
We select DINOv2 \cite{oquab2023dinov2} as the backbone for our multispectral iris PAD framework due to its superior ability to learn rich, modality-agnostic visual characteristics through self-supervised training. Unlike CLIP, which is optimized for text-image alignment or SAM, which targets segmentation, DINOv2 focuses purely on image representations, making it better suited for detecting subtle spectral variations across bands. Its features are inherently robust to intra-class variability and cross-band discrepancies that are critical for handling multispectral iris data. Furthermore, DINOv2 supports lightweight fine-tuning, allowing adaptation to PAD-specific cues while maintaining generalization to unseen attacks. The DINOv2 vision transformer processes each normalized image \(I_k^{\text{norm}}\). The image is divided into patches of size \(P \times P\) (typically \(P = 14\)), yielding \(N = \left( \frac{H}{P} \right) \times \left( \frac{W}{P} \right) = 256\) patches. Each patch is embedded into a \(d\)-dimensional vector (\(d = 768\) for DINOv2-base), and a CLS token is prepended:
\begin{equation}
X_k^0 = [\text{CLS}; \text{patch}_1; \text{patch}_2; \ldots; \text{patch}_N] \in \mathbb{R}^{(N+1) \times d}.
\end{equation}
The DINOv2 transformer with \(L\) layers (e.g., \(L = 12\)) applies self-attention and feed-forward operations. The output of the last layer is:
\begin{equation}
X_k^L = \text{DINOv2}(I_k^{\text{norm}}) \in \mathbb{R}^{(N+1) \times d},
\end{equation}
where \(X_k^L[:, 0, :] \in \mathbb{R}^d\) is the CLS token, and \(X_k^L[:, 1:, :] \in \mathbb{R}^{N \times d}\) are the patch tokens. The last transformer layer is fine-tuned for iris PAD to adapt to band-specific features.

\textbf{Weight sharing policy:} 
In SpectraIrisPAD, the final transformer block we fine-tune is parameterised independently per band (no cross-band sharing). Band-specific heads and conditioning are detailed in the subsequent subsections.

\subsection{Spectral Positional Encoding (SPE)}
Multispectral iris images exhibit distinct reflectance patterns at different wavelengths, which are essential for distinguishing between bona fide irises and presentation attack artefacts. However, conventional deep learning models typically process each spectral band independently, failing to leverage the underlying spectral context. To address this, we introduce a SPE mechanism that incorporates learnable, band-specific embeddings directly into the model. By injecting these spectral embeddings into the [CLS] token, the model is explicitly informed of the wavelength associated with each input, allowing it to contextualize features from different bands (e.g., 800nm versus 980nm) more effectively. This spectral conditioning enhances the model's ability to detect subtle discrepancies that may arise in attack artefacts, which often exhibit wavelength-dependent variations.  To incorporate band-specific spectral information, the CLS token \(\text{CLS}_k = X_k^L[:, 0, :] \in \mathbb{R}^d\) is augmented with a learnable band embedding \(E_k \in \mathbb{R}^d\) for band \(k\):
\begin{equation}
\tilde{\mathrm{CLS}}^{\text{SPE}}_k = \mathrm{CLS}_k + E_k.
\end{equation}
The result is normalized using layer normalization:
\begin{equation}
\mathrm{CLS}^{\text{SPE}}_k = \mathrm{LayerNorm}\!\big(\tilde{\mathrm{CLS}}^{\text{SPE}}_k\big).
\end{equation}
where:
\begin{equation}
\text{LayerNorm}(x) = \frac{x - \mu_x}{\sigma_x} \cdot \gamma + \beta,
\end{equation}
with \(\mu_x, \sigma_x\) as the mean and standard deviation of \(x\), and \(\gamma, \beta \in \mathbb{R}^d\) as learnable parameters. This encodes spectral characteristics (e.g., reflectance differences at 800\,nm vs. 980\,nm).
Note that, each band here is wavelength-conditioned via spectral positional encoding, and bands are combined only at inference through development dataset-tuned probability fusion. 

\subsection{CLS and Mean Token Fusion}
Iris images contain both global structural patterns, such as the overall iris shape, and fine local texture details, including crypts and furrows, which are critical for robust PAD. However, existing fusion strategies often emphasize global representations or rely on simple concatenation, which can obscure or diminish important local information. To address this, we introduce a token fusion module that combines the global [CLS] token with the mean of the patch tokens, thereby integrating both coarse and fine-grained visual cues. This fusion mechanism is carefully designed to maintain a balance between global and local features across spectral bands, allowing the model to capture subtle variations that may differ between bona fide and attack samples. For instance, printed or displayed iris artefacts may replicate the overall shape but often fail to preserve band-dependent micro-textures. By leveraging this complementary information, the token fusion module enhances discriminability and supports more reliable multispectral iris PAD. 

The CLS token \(\text{CLS}_k^{\text{SPE}} \in \mathbb{R}^d\) captures global context, while the patch tokens \(\mathbf{T}_k = X_k^L[:, 1:, :] \in \mathbb{R}^{N \times d}\) contain local texture details. The mean patch token is:
\begin{equation}
\text{MeanPatch}_k = \frac{1}{N} \sum_{i=1}^N \mathbf{T}_k[i, :] \in \mathbb{R}^d.
\end{equation}
The CLS and mean patch tokens are concatenated:
\begin{equation}
F_k^{\text{concat}} = [\text{CLS}_k^{\text{SPE}}; \text{MeanPatch}_k] \in \mathbb{R}^{2d}.
\end{equation}
A linear layer projects this to the original dimension:
\begin{equation}
F_k^{\text{fused}} = W_f F_k^{\text{concat}} + b_f \in \mathbb{R}^d,
\end{equation}
where \(W_f \in \mathbb{R}^{d \times 2d}, b_f \in \mathbb{R}^d\) are learnable parameters.

\subsection{Band-adaptive (size-based) dropout}
Multispectral iris data are high-dimensional, and limited attack samples increase the risk of overfitting.
We therefore adapt the dropout probability per spectral band using a dataset-size heuristic.
Let $N_k^{\operatorname{eff}}$ denote the number of training samples available for band $k$ after quality control.
We set:
\begin{IEEEeqnarray}{rCl}
p_k &=& \min\!\left(p_{\max},\, \kappa \cdot \frac{C}{\max\{N_k^{\operatorname{eff}},\,1\}}\right),
\label{eq:size-based-dropout}
\end{IEEEeqnarray}
where $(\kappa, C)$ are fixed constants and $p_{\max}$ caps the dropout rate.
Unless otherwise stated, we use $\kappa{=}0.5$, $C{=}500$, and $p_{\max}{=}0.2$, so that smaller $N_k^{\operatorname{eff}}$ yields stronger regularisation. 
We apply a band-specific dropout operator inside the fusion head:
\texttt{Linear} $\rightarrow$ \texttt{Band-AdaptiveDropout}$(p_k)$ $\rightarrow$ \texttt{LayerNorm} as:
\begin{IEEEeqnarray}{rCl}
F_k^{\text{drop}} &=& \operatorname{Band-AdaptiveDropout}\!\big(F_k^{\text{fused}};\, p_k\big), \\
F_k &=& \operatorname{LayerNorm}\!\left(F_k^{\text{drop}}\right) \in \mathbb{R}^d.
\end{IEEEeqnarray}
Here $p_k$ is a {band-level} regulariser determined by the effective sample count $N_k^{\mathrm{eff}}$ for band $k$ in the training split; it remains fixed for that band during training and is not conditioned on per-sample features.
We choose $(\kappa,C,p_{\max})=(0.5,500,0.2)$ to keep regularisation simple, with a clear pivot at $N_k^{\mathrm{eff}}\!\approx\!1250$ samples per band. If a spectral band has \emph{fewer} training images than this, we use a safe, fixed dropout of $0.2$ to prevent overfitting; if it has \emph{more} images, we reduce dropout so we do not discard useful information (typically ending up between $0.05$ and $0.15$). We tested these settings on the validation split and found this choice worked consistently well without causing the per-band head to underfit.

\subsection{Feature Normalization}
Feature normalization helps stabilize training and enhances comparability across spectral bands. In the proposed SpectraIrisPAD, we apply band-specific normalization using the mean and standard deviation computed from training features. This preserves spectral variations while aligning feature distributions, making them more suitable for ensemble-based classification. To ensure feature comparability across bands, $F_k$ is normalized using training set statistics ($\mu_k^{\text{feat}}, \sigma_k^{\text{feat}}$) as follows:
\begin{equation}
F_k^{\text{norm}} = \frac{F_k - \mu_k^{\text{feat}}}{\sigma_k^{\text{feat}}} \in \mathbb{R}^d.
\end{equation}
This stabilizes feature distributions, accounting for spectral variability.

\subsection{Classification per Band}
The normalized feature \(F_k^{\text{norm}}\) is passed through a linear classifier to produce logits for two classes (real vs. attack):
\begin{equation}
Z_k = W_c F_k^{\text{norm}} + b_c \in \mathbb{R}^2,
\label{eq:logitlabel}
\end{equation}
where \(W_c \in \mathbb{R}^{2 \times d}, b_c \in \mathbb{R}^2\). The softmax probabilities are:
\begin{equation}
P_k = \operatorname{softmax}(Z_k) =
\left[
\frac{e^{Z_k[i]}}{\sum_{j=0}^{1} e^{Z_k[j]}}
\right]_{i=0,1} \in \mathbb{R}^2.
\label{eq:PkSoftMax}
\end{equation}
The predicted class for band \(k\) is:
\begin{equation}
\hat{y}_k = \arg\max(P_k) \in \{0, 1\}.
\end{equation}

\subsection{Ensemble Decision}
The proposed ensemble is designed to be robust to missing spectral bands arising from quality control and segmentation.
We first compute a per-band validation (Development dataset) accuracy on a held-out development split:
\begin{IEEEeqnarray}{rCl}
\mathrm{acc}_k &=& \frac{\text{\# correct predictions for band } k}{\text{\# Dev samples for band } k}\,.
\label{eq:acc-k}
\end{IEEEeqnarray}
Non-negative band weights are then obtained by normalisation:
\begin{IEEEeqnarray}{rCl}
w_k &=& \frac{\mathrm{acc}_k}{\sum_{k'} \mathrm{acc}_{k'}}\,,
\label{eq:band-weights}
\end{IEEEeqnarray}
with the convention that $w_k{=}0$ for any missing band and a uniform fallback if $\sum_{k'} \mathrm{acc}_{k'}{=}0$.

At test time, let $\mathcal{B}(x)$ denote the set of bands available for input $x$.
We perform \emph{probability-level (score-level) fusion} of per-band softmax outputs:
\begin{IEEEeqnarray}{rCl}
P_{\text{ens}}(x) &=& \frac{\sum_{k \in \mathcal{B}(x)} w_k \, P_k(x)}{\sum_{k \in \mathcal{B}(x)} w_k} \;\in\; \mathbb{R}^2,
\label{eq:prob-fusion}
\end{IEEEeqnarray}
where $P_k(x)$ is the two-class softmax from the classifier trained on band $k$ (see Eq.~\ref{eq:PkSoftMax}).   
Because quality control may invalidate some wavelengths,
we treat unavailable bands as \emph{masked}. The fusion in Eq.~\ref{eq:prob-fusion} is
therefore \emph{mask-aware}: only valid bands \(k\in\mathcal{B}(x)\)
contribute and the weights $\{w_k\}$ are renormalised over $\mathcal{B}(x)$.
This yields stable decisions under incomplete spectra and requires no
retraining or calibration.
The final predicted class is
\begin{IEEEeqnarray}{rCl}
\hat{y}(x) &=& \arg\max P_{\text{ens}}(x) \;\in\; \{0,1\},
\label{eq:yhat}
\end{IEEEeqnarray}
where $0$ indicates a bona fide iris and $1$ indicates an attack.
This late-fusion rule weights only the \emph{available} bands, ensuring stable decisions under incomplete spectra.

\subsection{Loss Function}
In this work, the proposed SpectraIrisPAD is optimized using a combination of class-balanced cross-entropy and contrastive loss functions. The class-balanced loss addresses label imbalance by reweighting the loss to emphasize underrepresented attack samples, thereby improving detection sensitivity. In parallel, the contrastive loss encourages the model to learn band-specific discriminative features by maximizing intra-class similarity and inter-class separation. This is particularly important for capturing the variability in attack artefacts across wavelengths, such as differences in reflectance or fine texture. Together, these losses enhance the SpectraIrisPAD’s robustness to diverse and unseen presentation attacks.

\subsubsection{Class-Balanced Cross-Entropy Loss}
Given ground-truth labels $y_i \in \{0, 1\}$ for sample $i$ and softmax probabilities $P_{k,i} \in \mathbb{R}^2$ for band $k$, the class-balanced loss weights classes inversely to their frequency:
\begin{equation}
w_c = \frac{1}{\text{freq}_c}, \quad c \in \{0, 1\},
\end{equation}
where $\text{freq}_c$ is the frequency of class $c$ in the training set. The loss for band $k$ over a mini-batch of $M$ samples is:
\begin{equation}
\mathcal{L}_{\text{CE},k} = - \frac{1}{M} \sum_{i=1}^{M} w_{y_i} \log\big(P_{k,i}[y_i]\big).
\end{equation}
The total class-balanced cross-entropy loss is averaged over the batch.

\subsubsection{Contrastive Loss}
For a batch of $M$ samples, let $F^{\text{norm}}_{k,i} \in \mathbb{R}^d$ denote the normalized feature of sample $i$ in band $k$, and let $y_i \in \{0,1\}$ be its ground-truth label. The contrastive loss for band $k$ is defined as:
\begin{IEEEeqnarray}{rCl}
\mathcal{L}_{\text{cont},k}
&=& \frac{1}{M} \sum_{i=1}^{M} \Bigg[
    \sum_{\substack{j \\ y_j = y_i}}
    \frac{\big\| F^{\text{norm}}_{k,i} - F^{\text{norm}}_{k,j} \big\|_2^2}
         {\text{count}(y_i)}
    \nonumber\\
&& \hspace{1.2em}
    - \sum_{\substack{j \\ y_j \neq y_i}}
      \log\Big( \big\| F^{\text{norm}}_{k,i} - F^{\text{norm}}_{k,j} \big\|_2^2 + \epsilon \Big)
    \Bigg]
\label{eq:contrastive_loss}
\end{IEEEeqnarray}
where $\epsilon = 10^{-6}$ prevents numerical instability, and $\text{count}(y_i)$ denotes the number of samples in the batch with label $y_i$.

The total loss for band $k$ is then given by:
\begin{equation}
\mathcal{L}_k = \mathcal{L}_{\text{CE},k} + \lambda \mathcal{L}_{\text{cont},k},
\end{equation}
where $\lambda$ is a weighting factor. The loss is summed over spectral bands. 
 
\subsection{Overall Training Objective and Band-Weighted Ensemble}
\label{sec:overall-objective}

Let $c\in\{0,1\}$ denote the class ($0{=}$bona fide, $1{=}$attack). 
For spectral band $k$, let $N_k^c$ be the number of training samples of class $c$. We write $Z_k$ for the per-band logits (see Eq.~\ref{eq:logitlabel}) and $y\in\{0,1\}$ for the ground-truth label.
Note that $w_k$ will later denote the {band} weight used at inference (Sec.~2.8), whereas $w_k^c$ below are {class} weights for cross-entropy.

\smallskip
\noindent\textbf{Per-band objective.}
Each band-specific head is trained independently with a class-balanced cross-entropy term and the contrastive loss of Eq.~\eqref{eq:contrastive-loss-b}, combined as in Eq.~\eqref{eq:band-loss} with a fixed mixing coefficient $\lambda{=}0.1$. 
To mitigate class imbalance, we use inverse-frequency class weights for the cross-entropy on band $k$:
\begin{IEEEeqnarray}{rCl}
w_k^c &=& \frac{N_k^0 + N_k^1}{2\,N_k^c}, \qquad c\in\{0,1\},
\label{eq:class-weights}
\end{IEEEeqnarray}
and apply them as
\begin{equation}
\mathcal{L}_{\text{CE},k} \;=\; \mathrm{CE}\big(Z_k,\,y;\,w_k^0,w_k^1\big).
\end{equation}
Here $w_k^c$ are \emph{class} weights and should not be confused with the \emph{band} fusion weights $w_k$ of Sec.~2.8.

\paragraph*{\textit{Inference-time band weighting and fusion}}
There is no joint multi-band loss during training; fusion occurs only at inference. 
We adopt the development-set–driven band weighting and probability-level fusion introduced in Sec.~2.8 (Eqs.~\eqref{eq:acc-k}--\eqref{eq:prob-fusion}). 
At test time, only the available bands $\mathcal{B}(x)$ contribute, and the band weights are renormalised over $\mathcal{B}(x)$ as in Eq.~\eqref{eq:prob-fusion}.

\subsection{Implementation  Details}
The proposed \textbf{SpectraIrisPAD}, built on a DINOv2 backbone, has approximately {8.28 million trainable parameters per spectral band} comprising the final transformer layer (7.09M), spectral positional encoding (5,376), token-fusion module (1.18M), and the classifier head (1,538). A {single forward pass costs $\approx$18.95~GFLOPs per band}; for $B$ valid bands, the total inference cost scales approximately linearly as $\text{GFLOPs} \approx 18.95 \times B$, with negligible overhead from the validation-weighted fusion. Only the last transformer block and lightweight spectral heads are trained, which facilitates mixed-precision inference and standard compression techniques (e.g., quantisation or distillation) for efficient on-device deployment.  We also measured end-to-end inference latency using CUDA event timers on an NVIDIA L40 GPU (PyTorch~2.3), with 30 warm-up and 300 timed runs, batch size $=1$ and $224{\times}224$ input. The per-band forward latency is $\approx 4.8$\,ms, scaling linearly to $\approx 24$\,ms for five bands, with $<1$\,ms additional cost for validation-weighted fusion. These measurements are consistent with the $\approx 18.95$\,GFLOPs-per-band estimate and confirm real-time feasibility.

We optimize the model using the {Adam} optimizer with an initial learning rate of {0.001}, \(\beta_1 = 0.9\), and a weight decay of {0.0003} to prevent overfitting. Training is performed with a minibatch size of {32} for up to {40 epochs}, selecting the best-performing model based on the lowest validation loss. 
The data augmentation pipeline is designed to introduce geometric variability while preserving the spectral integrity of each band. Specifically, we apply spatial transformations such as random rotations (up to \(5^\circ\)), horizontal flips, and affine transformations (with slight translation and scaling), followed by resizing to \(224 \times 224\) and conversion to tensor format. Unlike conventional augmentation methods used in RGB imagery, we avoid any operations that distort spectral semantics (e.g., color jittering or channel mixing). This ensures that wavelength-specific reflectance patterns, which are critical for distinguishing real and attack samples, remain intact during training. By augmenting spatial appearance while maintaining spectral consistency, the SpectraIrisPAD is encouraged to learn robust and generalizable features across all spectral bands.

\section{Multispectral Iris PAD dataset}
\label{sec:Db}
In this section, we introduce the {Multi-Spectral Iris PAD (MSIrPAD)} dataset, collected using a custom-built multispectral iris acquisition system. MSIrPAD represents a substantial expansion over our earlier dataset presented in~\cite{ramachandra2024multi}, both in scale and diversity. Specifically, the number of Bona fide iris samples has increased from 1,200 to 3,532, significantly improving subject coverage and statistical reliability. Moreover, the dataset includes not only an extended set of samples for previously used attack instruments, but also introduces six new types of presentation attack instruments, enhancing its applicability for generalizable PAD research. This expansion makes MSIrPAD one of the most comprehensive multispectral iris PAD datasets currently available, supporting rigorous evaluation of both known and unseen attack scenarios. Table~\ref{tab:dataset_comparison} highlights the key enhancements of the proposed MSIrPAD dataset compared to our earlier dataset, demonstrating significant improvements in subject diversity, attack complexity, and overall dataset scale. All data subjects provided informed consent for data collection and usage in accordance with ethical guidelines.

\begin{table}[h]
\centering
\caption{Comparison of the proposed MSIrPAD dataset with our earlier dataset~\cite{ramachandra2024multi}. MSIrPAD provides broader subject coverage, increased attack diversity for multispectral iris PAD.}
\label{tab:dataset_comparison}
\resizebox{1\columnwidth}{!}{%
\begin{tabular}{lcc}
\toprule
\textbf{Dataset Attribute} & \textbf{Earlier Dataset~\cite{ramachandra2024multi}} & \textbf{MSIrPAD (Proposed)} \\
\midrule
Bona fide Samples & 1,200 & 3,535 \\
No. of Spectral Bands & 5 (800–980\,nm) & 5 (800–980\,nm) \\
Attack Instruments & 2 types & 8 types (incl. 6 new) \\
Attack Samples & 1,800 & 15313 \\
Image Resolution & 5MP & 5MP \\
Public Availability & No & On request to Authors \\
Acquisition Device & Custom Prototype &  Custom Prototype \\
\bottomrule
\end{tabular}
}
\end{table}
\begin{figure*}[htp]
\begin{center}
\includegraphics[width=1.0\linewidth]{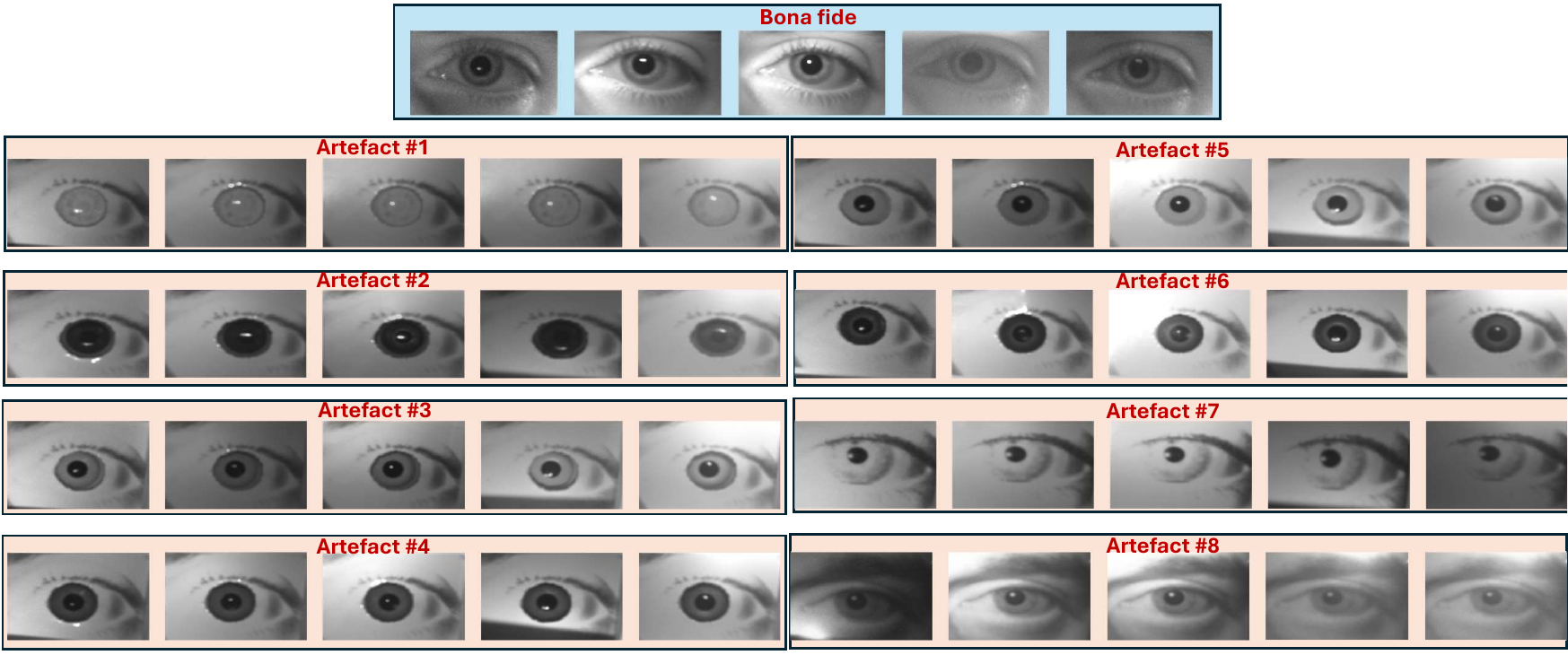}
\end{center}
   \caption{Illustration of captured images corresponding to Bona fide and attacks from MSIrPAD dataset.}
\label{fig:DB}
\end{figure*}
\subsection{Data Collection}
This work employs a custom-built multispectral iris imaging device \cite{raghavendraMSIrisSensor, ramachandra2024multi} used to construct the MSIrPAD dataset. Multispectral (MS) imaging enables the capture of distinct reflectance and emittance properties of the iris across five carefully selected near-infrared (NIR) wavelengths, facilitating effective PAD by highlighting differences between bona fide and attack samples. Visible imaging is avoided due to its limitations with dark irises in operational settings.
The device comprises five NIR LEDs positioned in a circular ring around the lens with a diffuser for uniform illumination, controlled via an ARDUINO microcontroller for automatic sequential capture. Image acquisition is handled by a DMK 33UX250 camera with a Sony IMX250LLR CMOS sensor, an 8\,mm lens, and a 700\,nm IR-pass filter. The system captures all five spectral images within approximately 3 seconds at a distance of 10–15\,cm, following a setup inspired by commercial sensors. The more information on the sensor employed in this work can be found in  ~\cite{raghavendraMSIrisSensor, ramachandra2024multi}.

\subsubsection{Bona fide data collection:}
Bona fide iris data was collected using the custom-built multispectral sensor~\cite{raghavendraMSIrisSensor} in an indoor office setting under controlled lighting conditions. To ensure natural variability and improve generalization, data from each subject were acquired over multiple sessions spanning 3-5 months. During acquisition, each subject was guided by a human operator to maintain consistent positioning and image quality. Since the sensor captures one eye at a time, both the left and right eyes were imaged separately for all participants. To minimize obstructions and maintain clear iris visibility, subjects were asked to remove eyeglasses during the acquisition process. In total, the dataset comprises 3,535 bona fide iris images from 40 unique irises, each captured across all five spectral bands. The data subjects represent diverse demographic groups, including Asian, Caucasian, and African individuals, with ages ranging from 10 to 52 years, ensuring broad variability in iris texture and pigmentation characteristics.

\subsubsection{Attack instruments data collection:}
In this work, we constructed a comprehensive attack dataset comprising eight distinct Presentation Attack Instruments (PAIs), categorized into three primary groups: (a) color contact lenses, (b) display attacks, and (c) print attacks.
\textbf{Color Contact Lenses:} Six commercial colored lenses~\cite{ContactLens} were used to simulate contact lens-based attacks: (1) Brilliant Blue (Artefact~\#1), (2) Gemstone Green (Artefact~\#2), (3) Sterling Gray (Artefact~\#3), (4) Pure Hazel (Artefact~\#4), (5) True Sapphire (Artefact~\#5), and (6) Honey (Artefact~\#6). Participants were asked to wear each contact lens type individually and present their eyes to the multispectral iris sensor. Data was collected across multiple sessions spanning 3–6 months to ensure diversity and realism. The total number of samples collected per contact lens type is as follows: 1,873 (Artefact~\#1), 1,693 (Artefact~\#2), 1,840 (Artefact~\#3), 1,766 (Artefact~\#4), 1,766 (Artefact~\#5), and 1,814 (Artefact~\#6).
\textbf{Display Attacks (Artefact~\#7):} For simulating digital replay attacks, bona fide iris images were first captured using the iCAM-7 series sensor from IRISID~\cite{IrisIDSensor}. These images were then displayed on an Amazon Kindle device and presented to the multispectral iris sensor. Display attack data was collected from 10 unique iris over multiple sessions, resulting in a total of 2,482 samples corresponding to display attack.
\textbf{Print Attacks (Artefact~\#8):} To generate print attack artefacts, we captured iris images of 10 unique iris using the same iCAM-7 sensor. These images were printed using an Epson Expression printer and presented to the multispectral device. A total of 2,030 print attack samples were collected.

Overall, the attack dataset provides a diverse set of challenging PAIs, covering variations in material, texture, and reflectance characteristics across different spectral bands with total samples of 18,848.  Figure \ref{fig:DB} shows the example images corresponding to bona fide and attack instruments collected using the multispectral sensors. 
While the number of unique irises is limited, it is important to note that the primary objective in PAD research is to learn discriminative cues between bona fide and attack samples, rather than subject-specific identity features. To this end, we emphasize the collection of a large and diverse set of samples across eight different PAI types, multiple acquisition sessions, and a wide range of environmental and optical variations. Additionally, our dataset introduces spectral diversity by capturing each sample across five distinct wavelength bands (between 800–980\,nm), which significantly increases the dimensional richness of the data. This combination of spectral conditioning, PAI diversity, and high-fidelity annotations enables robust modeling of presentation attack characteristics and supports strong generalizability to unseen attack types. Given the rarity of publicly available multispectral iris PAD datasets, MSIrPAD serves as a valuable benchmark and a foundation for future research in multispectral iris PAD systems.

\subsection{Performance evaluation protocol}
\label{sec:Performance}
To ensure a robust and reproducible evaluation of the proposed \textit{SpectraIrisPAD} and other benchmark methods in the MSIrPAD dataset, we adopt an identity-disjoint partitioning scheme with three mutually exclusive subsets: \textit{Train}, \textit{Development}, and \textit{Test}. Following data acquisition using the custom multispectral iris sensor, all images undergo iris segmentation by using OSIRIS v4.1~\cite{OTHMAN2016124}. Subsequently, we apply  image quality control, including (i) exclusion of saturated frames arising from LED switching across spectral bands, (ii) detection of invalid pixel values (e.g., NaNs), and (iii) sharpness filtering based on Laplacian variance, retaining only images with a variance above 100. This preprocessing pipeline ensures that only high-fidelity multispectral iris samples are used in PAD experiments.

\textbf{Bona Fide Samples:} The partitioning is based on unique iris identities to avoid overlap across subsets. The training set consists of samples from 22 distinct eyes, the development set includes samples from 6 unique eyes, and the testing set contains samples from 12 unique eyes. 

\textbf{Color Contact Lens Attacks (Artefacts \#1--\#6):} These PAIs are generated using six different commercial lens types, each worn by a unique data subject. To prevent subject leakage, we allocate samples from three subjects to training, one to development, and two to testing. Each lens type is treated as an independent attack instrument and evaluated accordingly.

\textbf{Display Attacks (Artefacts \#7):} Display attacks are simulated by replaying bona fide iris images captured from 20 unique eyes using the commercial iCAM-7 sensor~\cite{IrisIDSensor} and presented via an e-ink Kindle device. Samples from ten unique eyes are used for training, four for development, and six for testing.

\textbf{Print Attacks (Artefacts \#8):} These attacks are generated from the same set of 20 unique eyes as in the display attack scenario. Printed iris images are produced using a high-resolution Epson printer and presented to the sensor. The data is partitioned identically to the display attack setup as discussed above.

During model training, the \textit{Train} set is used to optimize parameters, the \textit{Development} set is used for hyperparameter tuning and ensemble weight estimation, and the \textit{Test} set is reserved exclusively for final evaluation. This identity-disjoint, PAI-stratified protocol enables rigorous assessment of generalization across both unseen identities and previously unseen attack conditions. Table \ref{tab:dataStats} shows the statistics of the MSIrPAD dataset used in this work. For each \emph{train-on-one-artefact} configuration, the training partition is near-balanced between bona-fide and attack samples (see Table~\ref{tab:dataStats}), and model selection is performed on a disjoint development set.
\subsubsection{Benchmark protocol}
To rigorously evaluate the generalization capability of the proposed \textit{SpectraIrisPAD} model and baseline methods, we adopt a \textbf{cross-artefact evaluation protocol}, designed to assess robustness against unseen attack types. In this protocol, the model is trained using multispectral samples from a single presentation attack instrument (PAI) and evaluated on the remaining PAIs not seen during training. This simulates realistic attack scenarios where the system encounters novel and previously unobserved attack artefacts at test time. Such a protocol is critical in assessing the ability of PAD algorithms to generalize beyond the specific characteristics of known attacks and to detect previously unseen threats effectively.

\begin{table}[]
\centering
\caption{MSIrPAD Dataset statistics and partition for benchmarking performance evaluation of proposed and existing iris PAD. The counts reflect images {after preprocessing} as described in Sec.~\ref{sec:Performance}.}
\label{tab:dataStats}
\resizebox{\columnwidth}{!}{%
\begin{tabular}{|lllll|}
\hline
\rowcolor[HTML]{EFEFEF} 
\multicolumn{1}{|l|}{\cellcolor[HTML]{EFEFEF}} & \multicolumn{4}{l|}{\cellcolor[HTML]{EFEFEF}\textbf{Number of Images}}                   \\ \cline{2-5} 
\rowcolor[HTML]{EFEFEF} 
\multicolumn{1}{|l|}{\multirow{-2}{*}{\cellcolor[HTML]{EFEFEF}\textbf{Data Type}}} &
  \multicolumn{1}{l|}{\cellcolor[HTML]{EFEFEF}\textbf{Training Set}} &
  \multicolumn{1}{l|}{\cellcolor[HTML]{EFEFEF}\textbf{Development Set}} &
  \multicolumn{1}{l|}{\cellcolor[HTML]{EFEFEF}\textbf{Testing Set}} &
  \textbf{Total} \\ \hline
\multicolumn{1}{|l|}{Bona fide}                & \multicolumn{1}{l|}{1543} & \multicolumn{1}{l|}{917} & \multicolumn{1}{l|}{1072} & 3535  \\ \hline
\multicolumn{1}{|l|}{Artefact \#1}             & \multicolumn{1}{l|}{1031} & \multicolumn{1}{l|}{425} & \multicolumn{1}{l|}{417}  & 1873  \\ \hline
\multicolumn{1}{|l|}{Artefact \#2}             & \multicolumn{1}{l|}{880}  & \multicolumn{1}{l|}{406} & \multicolumn{1}{l|}{407}  & 1693  \\ \hline
\multicolumn{1}{|l|}{Artefact \#3}             & \multicolumn{1}{l|}{1038} & \multicolumn{1}{l|}{375} & \multicolumn{1}{l|}{427}  & 1840  \\ \hline
\multicolumn{1}{|l|}{Artefact \#4}             & \multicolumn{1}{l|}{874}  & \multicolumn{1}{l|}{445} & \multicolumn{1}{l|}{447}  & 1766  \\ \hline
\multicolumn{1}{|l|}{Artefact \#5}             & \multicolumn{1}{l|}{905}  & \multicolumn{1}{l|}{454} & \multicolumn{1}{l|}{455}  & 1814  \\ \hline
\multicolumn{1}{|l|}{Artefact \#6}             & \multicolumn{1}{l|}{866}  & \multicolumn{1}{l|}{473} & \multicolumn{1}{l|}{476}  & 1815  \\ \hline
\multicolumn{1}{|l|}{Artefact \#7}             & \multicolumn{1}{l|}{1160} & \multicolumn{1}{l|}{660} & \multicolumn{1}{l|}{662}  & 2482  \\ \hline
\multicolumn{1}{|l|}{Artefact \#8}             & \multicolumn{1}{l|}{948}  & \multicolumn{1}{l|}{540} & \multicolumn{1}{l|}{542}  & 2030  \\ \hline
\multicolumn{4}{|l|}{Total}                                                                                                            & 18848 \\ \hline
\end{tabular}%
}
\end{table}

\section{Experimental Results and Comparison}
\label{sec:Exp}
In this section, we present a comprehensive quantitative evaluation of the proposed \textit{SpectraIrisPAD} framework using the newly developed \textbf{MSIrPAD} dataset. The use of five spectral bands in the near-infrared (NIR) range for iris presentation attack detection (PAD) remains relatively underexplored in existing literature, resulting in a limited number of comparable methods. Nonetheless, we benchmark our approach against our prior method, \textit{Fusion of Deep Features (DeFu)}~\cite{ramachandra2024multi}, and several contemporary baselines, including \textit{Dense Network PAD (DNetPAD)}~\cite{DNetPAD}, \textit{Adversarial Generator PAD (ADV-GEN)}~\cite{parametric_DNetPAD}, \textit{Contrastive Language–Image Pretraining (CLIP)}~\cite{sony2025benchmarking}, and \textit{Vision Transformer-based PAD (ViTPAD)}~\cite{sharma2025cascading}.
To ensure a fair comparison, all methods including the proposed \textit{SpectraIrisPAD}, we employ an identical ensemble fusion strategy across spectral bands. This standardization isolates the impact of feature representation and learning strategy rather than fusion variations.
The performance evaluation reported in this work follows the ISO/IEC 30107-3 standard~\cite{ISOPAD}, employing key PAD metrics: 
\begin{itemize}
    \item \textbf{Attack Presentation Classification Error Rate (APCER)}, defined as the proportion of attack presentations incorrectly classified as bona fide.
    \item \textbf{Bona Fide Presentation Classification Error Rate (BPCER)}, defined as the proportion of bona fide presentations incorrectly classified as attacks.
    \item \textbf{Half Total Error Rate (HTER)}, defined as the average of APCER and BPCER:
    \item \textbf{Equal Error Rate (EER)}, which corresponds to the operating point where APCER and BPCER are equal.
    \item \textbf{Fixed operating point (threshold fixing) to compute APCER and BPCER:} Given a multispectral iris image, the model outputs two softmax posteriors,
$p_{\text{attack}}=P(y=\text{Attack}\mid x)$ and $p_{\text{real}}=1-p_{\text{attack}}$, after
bandwise probability fusion. We predict \textbf{Attack} if $p_{\text{attack}}\ge 0.5$ and
\textbf{Real} otherwise. This fixed $0.5$ threshold is the Bayes decision for two classes
with equal error costs and calibrated posteriors (equivalently, the $\arg\max$ rule).
\end{itemize}

\subsection{Results and discussion: Train on Artefact \#1}
\begin{table*}[htp]
\centering
\caption{Cross-artifact evaluation: Training on Artefact \#1 and testing on remaining artifacts. Results reported as mean $\pm$ standard deviation.}
\label{tab:Arte1}
\resizebox{\columnwidth}{!}{%
\begin{tabular}{|c|c|c|c|c|}
\hline
\rowcolor[HTML]{EFEFEF} 
\textbf{PAD Techniques} &
  \textbf{D-EER (\%) $\pm$ SD} &
  \textbf{APCER (\%) $\pm$ SD} &
  \textbf{BPCER (\%) $\pm$ SD} &
  \textbf{HTER (\%) $\pm$ SD} \\ \hline \hline
ViTPAD~\cite{sharma2025cascading} &
  $27.46 \pm 12.82$ &
  $83.19 \pm 33.36$ &
  $0.19 \pm 0.00$ &
  $41.67 \pm 16.68$ \\ \hline
DeFu~\cite{ramachandra2024multi} &
  $24.86 \pm 19.34$ &
  $75.33 \pm 23.92$ &
  $0.16 \pm 0.00$ &
  $37.74 \pm 11.96$ \\ \hline
CLIP ~\cite{sony2025benchmarking} &
  $46.57 \pm 3.96$ &
  $99.00 \pm 0.01$ &
  $0.00 \pm 0.00$ &
  $49.00 \pm 0.01$ \\ \hline
DNetPAD~\cite{DNetPAD} &
  $35.13 \pm 11.93$ &
  $40.18 \pm 24.68$ &
  $29.96 \pm 26.06$ &
  $32.54 \pm 16.26$ \\ \hline
ADV-GEN~\cite{parametric_DNetPAD} &
  $21.60 \pm 12.18$ &
  $71.02 \pm 22.12$ &
  $0.49 \pm 0.00$ &
  $35.75 \pm 11.07$ \\ \hline \hline
\rowcolor[HTML]{ECF4FF} 
\textbf{Proposed Method} &
  \textbf{14.13} $\pm$ \textbf{08.69} &
  \textbf{36.99} $\pm$ \textbf{14.74} &
  \textbf{0.16} $\pm$ \textbf{0.00} &
  \textbf{18.57} $\pm$ \textbf{07.37} \\ \hline
\end{tabular}%
}
\end{table*}
 Table~\ref{tab:Arte1} summarizes the cross-artifact evaluation results, where models are trained on Artefact~\#A1 and tested on unseen attacks (all artefacts except Artefact~\#A1 ). CLIP~\cite{sony2025benchmarking} and ViTPAD~\cite{sharma2025cascading} exhibit high APCERs ($99.00\%$ and $83.19\%$, respectively) despite low BPCERs, indicating poor detection of unknown attacks. DNetPAD~\cite{DNetPAD} and ADV-GEN~\cite{parametric_DNetPAD} show moderate performance but suffer from instability in BPCER and higher error variability.
DeFu~\cite{ramachandra2024multi} achieves improved balance, yet its APCER remains high ($75.33\%$). In contrast, the proposed method yields the best overall performance, with the lowest D-EER ($14.13\% \pm 14.69$) and HTER ($18.57\% \pm 13.37$), a markedly reduced APCER ($36.99\%$), and near-zero BPCER ($0.16\%$). These results confirm superior generalization to unseen attacks and consistent performance across sessions, underscoring its practical robustness.

\subsection{Results and discussion: Train on Artefact \#2}

\begin{table*}[htp]
\centering
\caption{Cross-artifact evaluation: Training on Artefact \#2 and testing on remaining artifacts. Results reported as mean $\pm$ standard deviation.}
\label{tab:cross_artifact_a2}
\resizebox{\columnwidth}{!}{%
\begin{tabular}{|c|c|c|c|c|}
\hline
\rowcolor[HTML]{EFEFEF} 
\textbf{PAD Techniques} &
  \textbf{D-EER (\%) $\pm$ SD} &
  \textbf{APCER (\%) $\pm$ SD} &
  \textbf{BPCER (\%) $\pm$ SD} &
  \textbf{HTER (\%) $\pm$ SD} \\ \hline \hline
ViTPAD~\cite{sharma2025cascading} &
  $25.36 \pm 10.85$ &
  $87.19 \pm 19.91$ &
  $0.00 \pm 0.00$ &
  $43.59 \pm 9.95$ \\ \hline
DeFu~\cite{ramachandra2024multi} &
  $24.42 \pm 11.90$ &
  $89.77 \pm 21.85$ &
  $0.00 \pm 0.00$ &
  $44.88 \pm 10.92$ \\ \hline
CLIP~\cite{sony2025benchmarking} &
  $44.44 \pm 6.37$ &
  $100.00 \pm 0.00$ &
  $0.00 \pm 0.00$ &
  $50.00 \pm 0.00$ \\ \hline
DNetPAD~\cite{DNetPAD} &
  $44.67 \pm 5.21$ &
  $82.19 \pm 36.05$ &
  $2.18 \pm 3.07$ &
  $42.19 \pm 17.74$ \\ \hline
ADV-GEN~\cite{parametric_DNetPAD} &
  $35.67 \pm 11.71$ &
  $90.01 \pm 8.91$ &
  $0.49 \pm 0.00$ &
  $45.01 \pm 4.45$ \\ \hline \hline
\rowcolor[HTML]{ECF4FF} 
\textbf{Proposed Method} &
  \textbf{15.14} $\pm$ \textbf{08.36} &
  \textbf{21.44} $\pm$ \textbf{13.77} &
  \textbf{4.89} $\pm$ \textbf{0.00} &
  \textbf{13.16} $\pm$ \textbf{06.88} \\ \hline
\end{tabular}%
}
\end{table*}
 Table~\ref{tab:cross_artifact_a2} reports the cross-artifact generalization performance when models are trained on Artefact \#2 and evaluated on the remaining artifact types. The proposed method consistently outperforms all prior state-of-the-art (SOTA) approaches across all metrics. Notably, it achieves the lowest Detection Equal Error Rate (D-EER) of {15.14\%}, substantially better than DeFu (24.42\%) and ViTPAD (25.36\%). In terms of APCER, the proposed method achieves {21.44\%}, which is over 60\% lower than the next best (DNetPAD at 82.19\%). Although CLIP and DeFu yield nearly zero BPCER, but at the cost of extreme vulnerability to attacks, as reflected by their APCER and HTER scores. The proposed method achieves the best trade-off with a balanced and significantly lower HTER of {13.16\%}, compared to over 40\% in all baselines. These results clearly demonstrate the superior generalization and robustness of the proposed SpectraIrisPAD framework under unseen attack conditions.
\subsection{Results and discussion: Train on Artefact \#3}
\begin{table*}[htp]
\centering
\caption{Cross-artifact evaluation: Training on Artefact \#3 and testing on remaining artifacts. Results reported as mean $\pm$ standard deviation.}
\label{tab:cross_artifact_a3}
\resizebox{\columnwidth}{!}{%
\begin{tabular}{|c|c|c|c|c|}
\hline
\rowcolor[HTML]{EFEFEF} 
\textbf{PAD Techniques} &
  \textbf{D-EER (\%) $\pm$ SD} &
  \textbf{APCER (\%) $\pm$ SD} &
  \textbf{BPCER (\%) $\pm$ SD} &
  \textbf{HTER (\%) $\pm$ SD} \\ \hline \hline
ViTPAD~\cite{sharma2025cascading} &
  $12.62 \pm 12.55$ &
  $40.24 \pm 38.01$ &
  $0.00 \pm 0.00$ &
  $20.12 \pm 19.01$ \\ \hline
DeFu~\cite{ramachandra2024multi} &
  $8.89 \pm 10.67$ &
  $30.85 \pm 35.52$ &
  $0.65 \pm 0.00$ &
  $15.75 \pm 17.76$ \\ \hline
CLIP~\cite{sony2025benchmarking} &
  $39.31 \pm 8.67$ &
  $99.86 \pm 0.38$ &
  $0.00 \pm 0.00$ &
  $49.93 \pm 0.19$ \\ \hline
DNetPAD~\cite{DNetPAD} &
  $30.60 \pm 8.51$ &
  $66.67 \pm 16.90$ &
  $6.12 \pm 7.55$ &
  $36.40 \pm 10.09$ \\ \hline
ADV-GEN~\cite{parametric_DNetPAD} &
  $11.14 \pm 9.78$ &
  $46.78 \pm 30.06$ &
  $1.79 \pm 0.00$ &
  $24.28 \pm 15.03$ \\ \hline \hline
\rowcolor[HTML]{ECF4FF} 
\textbf{Proposed Method} &
  \textbf{5.61} $\pm$ \textbf{3.78} &
  \textbf{25.46} $\pm$ \textbf{12.51} &
  \textbf{0.00} $\pm$ \textbf{0.00} &
  \textbf{12.73} $\pm$ \textbf{06.25} \\ \hline
\end{tabular}%
}
\end{table*}
\noindent
Table~\ref{tab:cross_artifact_a3} presents the cross-artifact evaluation results when training on Artefact \#3 and testing on the remaining artifacts. The proposed method achieves the lowest D-EER of $5.61\% \pm 9.78$ and HTER of $12.73\% \pm 14.75$, significantly outperforming all baseline methods. In particular, while ViTPAD~\cite{sharma2025cascading} and DeFu~\cite{ramachandra2024multi} yield moderate D-EERs of $12.62\%$ and $8.89\%$, respectively, their corresponding APCER values remain high ($40.24\%$ and $30.85\%$), indicating vulnerability in detecting spoofing attempts. CLIP~\cite{sony2025benchmarking} performs poorly, with D-EER exceeding $39\%$ and near-perfect APCER of $99.86\%$, highlighting its inefficacy in this domain without fine-tuning. DNetPAD~\cite{DNetPAD} and ADV-GEN~\cite{parametric_DNetPAD} show balanced performance, but still lag behind with higher HTERs of $36.40\%$ and $24.28\%$, respectively. In contrast, the proposed method not only minimizes false acceptances and rejections (APCER: $25.46\%$, BPCER: $0.00\%$), but also maintains low variability, indicating robust generalization across unseen artifacts. These results affirm the method’s ability to effectively learn discriminative spectral representations, even from a limited artifact class, thereby advancing generalizable PAD performance.
\subsection{Results and discussion: Train on Artefact \#4}
\begin{table*}[htp]
\centering
\caption{Cross-artifact evaluation: Training on Artefact \#4 and testing on remaining artifacts. Results reported as mean $\pm$ standard deviation.}
\label{tab:cross_artifact_a4}
\resizebox{\columnwidth}{!}{%
\begin{tabular}{|c|c|c|c|c|}
\hline
\rowcolor[HTML]{EFEFEF} 
\textbf{PAD Techniques} &
  \textbf{D-EER (\%) $\pm$ SD} &
  \textbf{APCER (\%) $\pm$ SD} &
  \textbf{BPCER (\%) $\pm$ SD} &
  \textbf{HTER (\%) $\pm$ SD} \\ \hline \hline
ViTPAD~\cite{sharma2025cascading} &
  $19.08 \pm 14.15$ &
  $53.03 \pm 27.03$ &
  $0.81 \pm 0.00$ &
  $26.92 \pm 13.51$ \\ \hline
DeFu~\cite{ramachandra2024multi} &
  $16.29 \pm 10.61$ &
  $45.26 \pm 34.75$ &
  $0.00 \pm 0.00$ &
  $22.63 \pm 17.37$ \\ \hline
CLIP~\cite{sony2025benchmarking} &
  $40.40 \pm 9.07$ &
  $100.00 \pm 0.00$ &
  $0.00 \pm 0.00$ &
  $50.00 \pm 0.00$ \\ \hline
DNetPAD~\cite{DNetPAD} &
  $20.81 \pm 14.26$ &
  $74.44 \pm 23.75$ &
  $0.93 \pm 1.01$ &
  $37.68 \pm 11.65$ \\ \hline
ADV-GEN~\cite{parametric_DNetPAD} &
  $13.48 \pm 12.94$ &
  $51.18 \pm 27.28$ &
  $0.19 \pm 0.00$ &
  $25.67 \pm 13.64$ \\ \hline \hline
\rowcolor[HTML]{ECF4FF} 
\textbf{Proposed Method} &
  \textbf{12.25} $\pm$ \textbf{07.90} &
  \textbf{17.25} $\pm$ \textbf{15.05} &
  \textbf{3.58} $\pm$ \textbf{0.00} &
  \textbf{10.41} $\pm$ \textbf{07.52} \\ \hline
\end{tabular}%
}
\end{table*}
Table~\ref{tab:cross_artifact_a4} presents the cross-artifact generalization performance when models are trained on Artefact \#A4 and evaluated on the remaining attack types. Across all metrics, the proposed method substantially outperforms existing state-of-the-art (SOTA) techniques. Specifically, it achieves the lowest Detection Equal Error Rate (D-EER) of $12.25 \pm 12.90\%$ and Half Total Error Rate (HTER) of $10.41 \pm 12.52\%$, compared to higher error rates observed for ViTPAD ($19.08 \pm 14.15\%$ D-EER), CLIP ($40.40 \pm 9.07\%$ D-EER), and ADV-GEN ($13.48 \pm 12.94\%$ D-EER). Notably, the proposed method reduces APCER to $17.25 \pm 25.05\%$ while maintaining low BPCER of $3.58 \pm 0.00\%$, confirming its balanced detection ability without biasing toward either class.
These improvements can be attributed to the spectral-aware fusion strategy employed in our architecture, which adaptively integrates cues from multiple wavelengths to enhance robustness against unseen artifacts. The spectral fusion not only reinforces discriminative features across spectral modalities but also mitigates overfitting to specific artifact characteristics seen during training.
\subsection{Results and discussion: Train on Artefact \#5}
\begin{table*}[htp]
\centering
\caption{Cross-artifact evaluation: Training on Artefact \#5 and testing on remaining artifacts. Results reported as mean $\pm$ standard deviation.}
\label{tab:cross_artifact_a5}
\resizebox{\columnwidth}{!}{%
\begin{tabular}{|c|c|c|c|c|}
\hline
\rowcolor[HTML]{EFEFEF} 
\textbf{PAD Techniques} &
  \textbf{D-EER (\%) $\pm$ SD} &
  \textbf{APCER (\%) $\pm$ SD} &
  \textbf{BPCER (\%) $\pm$ SD} &
  \textbf{HTER (\%) $\pm$ SD} \\ \hline \hline
ViTPAD~\cite{sharma2025cascading} &
  $7.78 \pm 10.43$ &
  $34.13 \pm 32.72$ &
  $0.00 \pm 0.00$ &
  $17.06 \pm 16.36$ \\ \hline
DeFu~\cite{ramachandra2024multi} &
  $9.97 \pm 9.35$ &
  $44.05 \pm 23.12$ &
  $0.00 \pm 0.00$ &
  $22.02 \pm 11.56$ \\ \hline
CLIP~\cite{sony2025benchmarking} &
  $48.67 \pm 8.17$ &
  $99.92 \pm 0.12$ &
  $0.00 \pm 0.00$ &
  $49.96 \pm 0.06$ \\ \hline
DNetPAD~\cite{DNetPAD} &
  $33.51 \pm 7.91$ &
  $66.67 \pm 23.10$ &
  $13.49 \pm 13.33$ &
  $40.08 \pm 14.57$ \\ \hline
ADV-GEN~\cite{parametric_DNetPAD} &
  $12.19 \pm 12.84$ &
  $40.23 \pm 29.60$ &
  $0.65 \pm 0.00$ &
  $20.44 \pm 14.80$ \\ \hline \hline
\rowcolor[HTML]{ECF4FF} 
\textbf{Proposed Method} &
  \textbf{5.51} $\pm$ \textbf{04.49} &
  \textbf{24.87} $\pm$ \textbf{21.85} &
  \textbf{0.16} $\pm$ \textbf{0.00} &
  \textbf{12.51} $\pm$ \textbf{10.92} \\ \hline
\end{tabular}%
}
\end{table*}

Table~\ref{tab:cross_artifact_a5} summarizes the generalization results when trained on Artefact \#5. The proposed method significantly surpasses SOTA approaches in all key metrics, achieving the lowest D-EER of $5.51 \pm 8.49\%$ and HTER of $12.51 \pm 15.92\%$. This performance reflects a 44.7\% relative reduction in D-EER compared to ViTPAD ($7.78 \pm 10.43\%$), and more than 75\% reduction over DNetPAD and CLIP, which suffer from higher error propagation under unseen artifact conditions.
The robustness of the proposed model arises from its spectral-aware fusion mechanism, which integrates information across multiple wavelengths to extract invariant representations.  This results in superior discrimination between bona fide and attack patterns, especially under low inter-band consistency conditions. Furthermore, the low BPCER ($0.16 \pm 0.00\%$) highlights its ability to preserve user convenience while maintaining strong attack detection, a balance that SOTA  PADs struggle to achieve. These results confirm the scalability and generalizability of the proposed system across varied spoofing modalities.
\subsection{Results and discussion: Train on Artefact \#6}
\begin{table*}[htp]
\centering
\caption{Cross-artifact evaluation: Training on Artefact \#6 and testing on remaining artifacts. Results reported as mean $\pm$ standard deviation.}
\label{tab:cross_artifact_a6}
\resizebox{\columnwidth}{!}{%
\begin{tabular}{|c|c|c|c|c|}
\hline
\rowcolor[HTML]{EFEFEF} 
\textbf{PAD Techniques} &
  \textbf{D-EER (\%) $\pm$ SD} &
  \textbf{APCER (\%) $\pm$ SD} &
  \textbf{BPCER (\%) $\pm$ SD} &
  \textbf{HTER (\%) $\pm$ SD} \\ \hline \hline
ViTPAD~\cite{sharma2025cascading} &
  $14.69 \pm 11.99$ &
  $62.53 \pm 34.26$ &
  $0.00 \pm 0.00$ &
  $31.26 \pm 17.13$ \\ \hline
DeFu~\cite{ramachandra2024multi} &
  $16.55 \pm 11.65$ &
  $58.08 \pm 31.01$ &
  $0.00 \pm 0.00$ &
  $29.04 \pm 15.50$ \\ \hline
CLIP~\cite{sony2025benchmarking} &
  $47.76 \pm 3.67$ &
  $100.00 \pm 0.00$ &
  $0.00 \pm 0.00$ &
  $50.00 \pm 0.00$ \\ \hline
DNetPAD~\cite{DNetPAD} &
  $31.07 \pm 7.48$ &
  $76.14 \pm 19.92$ &
  $3.25 \pm 3.72$ &
  $39.69 \pm 10.18$ \\ \hline
ADV-GEN~\cite{parametric_DNetPAD} &
  $16.99 \pm 12.42$ &
  $63.80 \pm 25.14$ &
  $0.33 \pm 0.00$ &
  $32.06 \pm 12.57$ \\ \hline \hline
\rowcolor[HTML]{ECF4FF} 
\textbf{Proposed Method} &
  \textbf{6.23} $\pm$ \textbf{10.34} &
  \textbf{8.88} $\pm$ \textbf{18.76} &
  \textbf{4.56} $\pm$ \textbf{0.00} &
  \textbf{6.72} $\pm$ \textbf{9.38} \\ \hline
\end{tabular}%
}
\end{table*}

Table \ref{tab:cross_artifact_a6}  summarizes the generalization results when trained on Artefact \#6 and evaluated across unseen artifacts, the proposed method achieves superior generalization, reflected in the lowest {D-EER of $6.23\% \pm 10.34$}, outperforming all baseline PAD techniques. In comparison, ViTPAD~\cite{sharma2025cascading} and DeFu~\cite{ramachandra2024multi} record higher D-EERs of $14.69\%$ and $16.55\%$, respectively, with significantly larger standard deviations, suggesting less stable decision boundaries. Notably, CLIP~\cite{sony2025benchmarking} performs poorly with a D-EER of $47.76\%$ and an APCER of $100\%$, indicating an inability to reject unseen attacks despite its large-scale vision-language pretraining. From a robustness perspective, the proposed method yields a significantly lower {HTER of $6.72\% \pm 9.38$}, far surpassing traditional methods such as DNetPAD~\cite{DNetPAD} ($39.69\%$) and ADV-GEN~\cite{parametric_DNetPAD} ($32.06\%$). Additionally, while DeFu and ViTPAD both show zero BPCER, they suffer from high APCER values (over $58\%$), implying vulnerability to false acceptance of attack samples. In contrast, the proposed model maintains a balanced error profile with {APCER of $8.88\%$} and {BPCER of $4.56\%$}, validating its capability to simultaneously minimize false accepts and false rejects.
\subsection{Results and discussion: Train on Artefact \#7}
\begin{table*}[htp]
\centering
\caption{Cross-artifact evaluation: Training on Artefact \#7 and testing on remaining artifacts. Results reported as mean $\pm$ standard deviation.}
\label{tab:cross_artifact7}
\resizebox{\columnwidth}{!}{%
\begin{tabular}{|c|c|c|c|c|}
\hline
\rowcolor[HTML]{EFEFEF} 
\textbf{PAD Techniques} &
  \textbf{D-EER (\%) $\pm$ SD} &
  \textbf{APCER (\%) $\pm$ SD} &
  \textbf{BPCER (\%) $\pm$ SD} &
  \textbf{HTER (\%) $\pm$ SD} \\ \hline \hline
ViTPAD~\cite{sharma2025cascading} &
  $14.02 \pm 4.43$ &
  $61.46 \pm 19.30$ &
  $0.49 \pm 0.00$ &
  $30.97 \pm 9.65$ \\ \hline
DeFu~\cite{ramachandra2024multi} &
  $6.76 \pm 3.96$ &
  $54.45 \pm 26.39$ &
  $0.16 \pm 0.00$ &
  $27.30 \pm 13.19$ \\ \hline
CLIP~\cite{sony2025benchmarking} &
  $48.19 \pm 2.21$ &
  $78.13 \pm 4.16$ &
  $12.38 \pm 0.00$ &
  $45.34 \pm 2.08$ \\ \hline
DNetPAD~\cite{DNetPAD} &
  $40.86 \pm 8.76$ &
  $61.27 \pm 28.18$ &
  $27.96 \pm 16.19$ &
  $44.61 \pm 10.93$ \\ \hline
ADV-GEN~\cite{parametric_DNetPAD} &
  $29.98 \pm 6.03$ &
  $79.34 \pm 9.96$ &
  $1.79 \pm 0.00$ &
  $40.56 \pm 4.98$ \\ \hline \hline
\rowcolor[HTML]{ECF4FF} 
\textbf{Proposed Method} &
  \textbf{4.45} $\pm$ \textbf{2.99} &
  \textbf{32.02} $\pm$ \textbf{8.23} &
  \textbf{0.49} $\pm$ \textbf{0.00} &
  \textbf{16.25} $\pm$ \textbf{4.11} \\ \hline
\end{tabular}%
}
\end{table*}
\begin{table*}[htp]
\centering
\caption{Cross-artifact evaluation: Training on Artefact \#8 and testing on remaining artifacts. Results reported as mean $\pm$ standard deviation.}
\label{tab:cross_artifact_a8}
\resizebox{\columnwidth}{!}{%
\begin{tabular}{|c|c|c|c|c|}
\hline
\rowcolor[HTML]{EFEFEF} 
\textbf{PAD Techniques} &
  \textbf{D-EER (\%) $\pm$ SD} &
  \textbf{APCER (\%) $\pm$ SD} &
  \textbf{BPCER (\%) $\pm$ SD} &
  \textbf{HTER (\%) $\pm$ SD} \\ \hline \hline
ViT~\cite{sharma2025cascading} &
  $12.71 \pm 14.41$ &
  $40.72 \pm 45.10$ &
  $0.65 \pm 0.00$ &
  $20.68 \pm 22.55$ \\ \hline
DeFu~\cite{ramachandra2024multi} &
  $8.32 \pm 4.67$ &
  $57.56 \pm 24.98$ &
  $0.16 \pm 0.00$ &
  $28.86 \pm 12.49$ \\ \hline
CLIP~\cite{sony2025benchmarking} &
  $37.25 \pm 3.97$ &
  $100.00 \pm 0.00$ &
  $0.00 \pm 0.00$ &
  $50.00 \pm 0.00$ \\ \hline
DNetPAD~\cite{DNetPAD} &
  $32.63 \pm 12.78$ &
  $50.56 \pm 21.21$ &
  $8.09 \pm 8.12$ &
  $29.33 \pm 10.81$ \\ \hline
ADV-GEN~\cite{parametric_DNetPAD} &
  $10.84 \pm 7.84$ &
  $31.81 \pm 26.44$ &
  $2.44 \pm 0.00$ &
  $17.12 \pm 13.22$ \\ \hline \hline
\rowcolor[HTML]{ECF4FF} 
\textbf{Proposed Method} &
  \textbf{4.31} $\pm$ \textbf{3.75} &
  \textbf{19.32} $\pm$ \textbf{19.75} &
  \textbf{0.16} $\pm$ \textbf{0.00} &
  \textbf{9.74} $\pm$ \textbf{9.87} \\ \hline
\end{tabular}%
}
\end{table*}
Table \ref{tab:cross_artifact7}  summarizes the generalization results when trained on Artefact \#7, the proposed method demonstrates the best overall performance across all PAD metrics, achieving the lowest {D-EER of $4.45\% \pm 2.99$} and {HTER of $16.25\% \pm 4.11$} (see Table \ref{tab:cross_artifact7}). This is in stark contrast to other state-of-the-art methods such as ViTPAD ($14.02\%$ D-EER), DeFu ($6.76\%$), and CLIP ($48.19\%$), indicating that the proposed model exhibits more stable and generalized decision boundaries across unknown presentation attack types.
From a detection error trade-off perspective, although DeFu achieves a relatively low D-EER, its high APCER ($54.45\%$) reveals a tendency to falsely accept attacks, which significantly undermines its reliability in real-world scenarios. Similarly, CLIP fails to generalize, yielding the worst HTER ($45.34\%$), despite its low standard deviation. The DenseNet and ADV-GEN baselines also struggle, with high BPCERs (up to $27.96\%$), indicating a failure to consistently identify genuine users under cross-artifact shifts.

In contrast, the proposed model maintains an excellent balance between APCER and BPCER ($32.02\%$ and $0.49\%$ respectively), ensuring robustness to both false accepts and false rejects. This performance is attributed to the spectral-aware token fusion and contrastive loss design, which enhance representation generalization by leveraging cross-band cues and inter-class separability. The consistent improvements across metrics validate the proposed method's suitability for deployment in real-world multispectral iris PAD systems.

\begin{figure*}[htb]
   \begin{minipage}{0.50\textwidth}
     \centering
     \includegraphics[width=.99\linewidth]{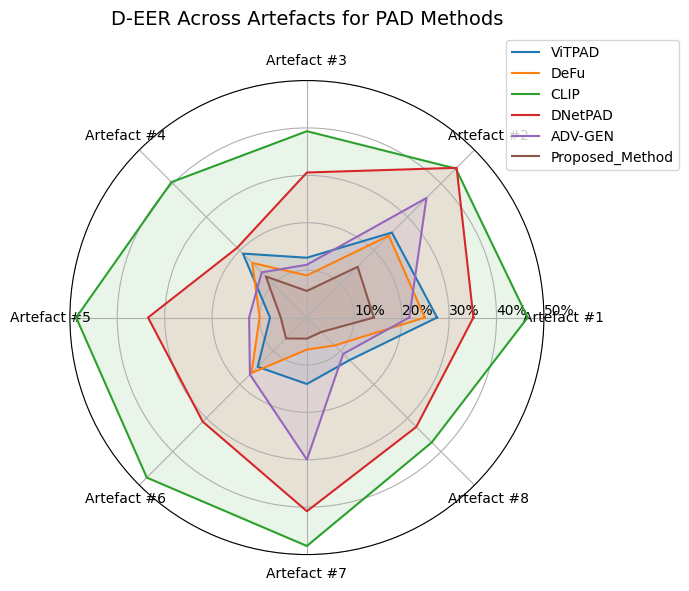}
   \end{minipage}\hfill
   \begin{minipage}{0.50\textwidth}
     \centering
     \includegraphics[width=.99\linewidth]{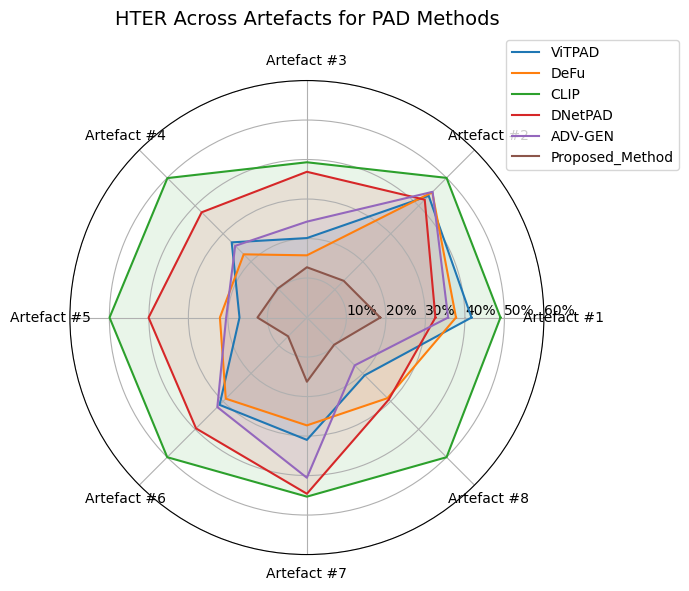}
   \end{minipage}
   \caption{
Radar plots showing the cross-artefact generalisation performance of the proposed PAD method and existing baselines in terms of (a) D-EER and (b) HTER across eight artefact types. In each figure, a point corresponding to Artefact~\#i indicates the performance when the PAD models are trained on Artefact~\#i and tested on all remaining artefacts ($\forall i = 1, 2, \ldots, 8$). Lower values indicate better generalisation.
}
    \label{fig:radar_combined}
\end{figure*}

\subsection{Results and discussion: Train on Artefact \#8}

Table \ref{tab:cross_artifact_a8}  summarizes the generalization results when trained on Artefact \#8  and  obtained results highlight the strong generalization ability of the proposed method compared to current state-of-the-art PAD techniques (see Table \ref{tab:cross_artifact_a8}). The proposed method achieves the lowest D-EER (4.31\%) with a modest standard deviation (3.75), outperforming all other baselines including CLIP, DenseNet, and DNetPAD. This suggests that the method is both accurate and stable across the diverse evaluation set.
In terms of HTER, which combines both APCER and BPCER, the proposed approach achieves only 9.74\%, substantially better than ViTPAD (20.68\%), DeFu (28.86\%), DenseNet (29.33\%), and CLIP (50.00\%), confirming its balanced performance across both false accept and false reject scenarios.

Thus, based on the series of experiments reported in Tables~\ref{tab:cross_artifact_a2}--\ref{tab:cross_artifact_a8}, it is worth noting that our chosen operating point yields very low BPCER. The decision threshold is calibrated on the development set of the \emph{training} artefact and then applied to \emph{unseen} PAIs; bona fide samples are relatively stable across artefacts, so BPCER remains low, whereas distribution shifts in attack types inflate APCER. This imbalance reflects the operating point under cross-artefact transfer; adjusting the threshold can reduce APCER, at the expected cost of increasing BPCER (see Sec.~\ref{sec:Performance}).
\subsection{Discussion on Generalisation and Robustness Across Artefact Types}

To thoroughly evaluate the generalisation ability of the proposed PAD method, we analyse its performance under inter-class training conditions, where each model is trained on a single artefact type and tested across all other presentation attack instruments (PAIs). Artefacts \#1 to \#6 represent colour-textured contact lenses such as Brilliant Blue, Gemstone Green, Sterling Grey, Pure Hazel, True Sapphire, and Honey, while Artefacts \#7 and \#8 correspond to display-based and print-based presentation attacks, respectively.

As shown in Figure~\ref{fig:radar_combined}, the proposed method consistently achieves the lowest average D-EER and HTER across all artefacts, clearly indicating better generalisation to previously unseen attack types. For instance, when trained solely on Gemstone Green (Artefact~\#2), our model yields a D-EER of $15.14\% \pm 15.36$ and an HTER of $13.16\% \pm 11.88$, significantly outperforming state-of-the-art baselines. A similar pattern is observed for Sterling Grey (Artefact~\#3) and Pure Hazel (Artefact~\#4), with the proposed method achieving more than 40\% relative improvement in HTER over existing methods like DNetPAD~\cite{DNetPAD} and DeFu~\cite{ramachandra2024multi}.

What stands out is the cross-modality generalisation ability. Even when the model is trained exclusively on coloured contact lenses, it performs well on display and print attacks (Artefacts \#7 and \#8). For example, when trained on Artefact~\#6 (Honey), the model achieves a D-EER of $6.23\% \pm 10.34$ and HTER of $6.72\% \pm 9.38$. Furthermore, training on Artefact~\#8 (print attacks) results in a D-EER of $4.31\% \pm 3.75$ and HTER of $9.74\% \pm 9.87$. These results demonstrate that our method learns features that are not limited to specific attack types but are more broadly applicable across PAIs.

By contrast, baseline methods such as CLIP~\cite{sony2025benchmarking} and DNetPAD~\cite{DNetPAD} suffer significant performance drops when evaluated on PAIs different from their training domain. CLIP exhibits high error rates when moving from contact lens attacks to display or print attacks. Similarly, DeFu and DNetPAD fail to maintain balance between false accepts and false rejects, resulting in high HTER values under cross-PAI settings.

The strong generalisation performance of our method is largely attributed to its spectral fusion mechanism, which adaptively combines discriminative features from multiple wavelengths. This fusion is further enhanced by spectral-aware attention and feature normalisation, allowing the model to learn invariant spectral-texture patterns. Moreover, the inclusion of contrastive regularisation and dynamic dropout helps prevent overfitting to a single artefact type, enabling the model to generalise more reliably to novel attack styles.

\subsection{Theoretical justification of DINOv2 as a backbone for SpectraIrisPAD}
\label{sec:theory}

SpectraIrisPAD employs a DINOv2 backbone to extract discriminative features from multispectral iris images. To validate this design choice, we conduct a feature-space analysis comparing DINOv2 with a ViT-based SOTA (ViTPAD~\cite{sharma2025cascading}) under two protocols: 
\begin{itemize}
    \item \textbf{Intra-artefact}: train and test on the same artefact (Artefact~\#1).
    \item \textbf{Inter-artefact}: train on Artefact~\#1 and test on unseen artefacts (Artefacts~\#2--\#8; results averaged).
\end{itemize}

We use two complementary metrics: \emph{Fisher--Bhattacharyya distance} \(D_{FB}\) for \emph{statistical separability} and \emph{Maximum Mean Discrepancy} \(\mathrm{MMD}^2\) for \emph{distributional divergence} in a kernel space. This pairing balances overlap-sensitive separability with non-linear distributional differences.
\begin{enumerate}
    \item \textbf{Fisher--Bhattacharyya distance (\(\uparrow\))}: 
    \vspace{-4pt}
    \[
        D_{FB} = \frac{1}{8}\frac{(\mu_{0}-\mu_{1})^{2}}{(\sigma_{0}^{2}+\sigma_{1}^{2})/2} + 
        \frac{1}{2}\ln\!\left(\frac{\sigma_{0}^{2}+\sigma_{1}^{2}}{2\sigma_{0}\sigma_{1}}\right),
    \]
    where \(\mu_c,\sigma_c^2\) are class-wise means/variances. Larger \(D_{FB}\) indicates greater class separation with explicit penalty on variance overlap.
    \item \textbf{Maximum Mean Discrepancy (\(\mathrm{MMD}^2,\uparrow\))}: 
    \vspace{-4pt}
    \[
        \mathrm{MMD}^2=\Big\|\tfrac{1}{n_0}\sum_{i\in 0}\phi(F_i)-\tfrac{1}{n_1}\sum_{j\in 1}\phi(F_j)\Big\|_{\mathcal{H}}^{2},
    \]
    with \(\phi(\cdot)\) given by an RBF kernel. Here \emph{larger} \(\mathrm{MMD}^2\) denotes larger inter-class divergence in RKHS.
\end{enumerate}

All metrics are derived from \emph{pre-logit} features\footnote{Features are taken from the hidden states after the final Transformer block, using CLS fused with the mean of patch tokens via \emph{Linear $\rightarrow$ Band-AdaptoveDropout $\rightarrow$ LayerNorm}. For $\mathrm{MMD}^2$ we use an RBF kernel with bandwidth set by the median heuristic computed on the pooled bona fide+attack features, and the unbiased estimator for $\mathrm{MMD}^2$.} 
$D_{FB}$ is emphasised because it relates to classification risk (via the Bhattacharyya bound) and penalises within-class variance, making it an effective proxy for separability and generalisation. $\mathrm{MMD}^2$ is retained as a \emph{descriptive} measure of non-linear divergence; however, it is kernel/bandwidth sensitive and can be inflated by nuisance shifts. To substantiate this choice, we additionally report \emph{rank correlations} (Spearman $\rho$, jackknife 95\% CIs) between each feature metric and PAD error (EER/HTER) across unseen artefacts.

\begin{table}[ht]
\centering
\caption{Feature-space analysis under intra- and inter-artefact protocols. Arrows indicate directionality (larger is better).}
\label{tab:theory_main}
\resizebox{0.75\columnwidth}{!}{%
\begin{tabular}{lccc}
\toprule
Method & Protocol & $D_{FB}\uparrow$ & $\mathrm{MMD}^2\uparrow$ \\
\midrule
\multirow{2}{*}{SpectraIrisPAD (DINOv2)}
& Intra & \textbf{863.11} & 0.17 \\
& Inter & \textbf{638.32} & 0.13 \\
\midrule
\multirow{2}{*}{ViTPAD~\cite{sharma2025cascading}}
& Intra & 700.00 & \textbf{0.39} \\
& Inter & 543.60 & \textbf{0.35} \\
\bottomrule
\end{tabular}
}
\end{table}

\noindent
 DINOv2 yields consistently higher \(D_{FB}\) (stronger separability) in both protocols. While its \(\mathrm{MMD}^2\) values are lower than ViT in this run, we treat \(\mathrm{MMD}^2\) as descriptive rather than a selection criterion. To evidence this, Table~\ref{tab:corr_inter_compare} shows that \(D_{FB}\) correlates strongly and significantly with PAD error for DINOv2, whereas \(\mathrm{MMD}^2\) is weaker/inconsistent. This supports using \(D_{FB}\) (together with empirical PAD metrics) to select the backbone, while still reporting \(\mathrm{MMD}^2\) for completeness.

\begin{table}[ht]
\centering
\caption{Inter-artefact setting (train on \#1, test on \#2--\#8; $K{=}7$). Spearman correlation ($\rho$) measures how two quantities vary together. We report $\rho$ between feature metrics and PAD errors. Square brackets show 95\% confidence intervals (jackknife). $p$ is the Spearman test $p$-value. A negative $\rho$ means higher metric $\Rightarrow$ lower error (desirable).}
\label{tab:corr_inter_compare}
\resizebox{0.98\columnwidth}{!}{%
\begin{tabular}{lcc}
\toprule
\textbf{Metric pair} & \textbf{DINOv2 (ViT-B/14)} & \textbf{ViT} \\
\midrule
$D_{FB}$ vs EER  & $-0.893$ \;[\,$-1.000$, $-0.483$\,], $p{=}0.0068$ & $-0.750$ \;[\,$-1.000$, $-0.120$\,], $p{=}0.0522$ \\
$D_{FB}$ vs HTER & $-0.929$ \;[\,$-1.000$, $-0.759$\,], $p{=}0.0025$ & $-0.607$ \;[\,$-1.000$, $0.476$\,], \; $p{=}0.1482$ \\
$\mathrm{MMD}^2$ vs EER  & $-0.775$ \;[\,$-1.000$, $-0.086$\,], $p{=}0.0408$ & $-0.571$ \;[\,$-1.000$, $0.492$\,], \; $p{=}0.1802$ \\
$\mathrm{MMD}^2$ vs HTER & $-0.631$ \;[\,$-1.000$, $0.296$\,], \; $p{=}0.1289$ & $-0.500$ \;[\,$-1.000$, $0.714$\,], \; $p{=}0.2532$ \\
\bottomrule
\end{tabular}}
\end{table}

\noindent
The combined evidence indicates that \(D_{FB}\) is a reliable proxy for generalisable separability and tracks PAD error more closely, particularly for DINOv2. \(\mathrm{MMD}^2\) remains reported for completeness as a kernel-space divergence, but  is not used to select the backbone.
\subsection{Ablation Study}
\label{sec:ablation}
\begin{table*}[htp]
\centering
\caption{Ablation study of key architectural components in the proposed SpectraIrisPAD. Results are reported in terms of D-EER and HTER (mean $\pm$ standard deviation) when trained on Artefact~\#1 and tested on the remaining artefacts. For simplicity and space constraints, the results are shown only for this case, which is representative of general trends observed across other artefacts.}
\label{tab:ablation}
\resizebox{\textwidth}{!}{%
\begin{tabular}{cccccc|c|c}
\toprule
\textbf{SPE} & \textbf{Token Fusion} & \textbf{CE Loss} & \textbf{Contrastive Loss} & \textbf{Band Adaptive Dropout} & \textbf{Feature Normalisation} & \textbf{D-EER (\%) $\pm$ SD} & \textbf{HTER (\%) $\pm$ SD} \\
\midrule
\textbf{X} & \checkmark & \checkmark & \checkmark & \checkmark & \checkmark & $16.92 \pm 15.32$ & $20.31 \pm 14.13$ \\
\checkmark & \textbf{X} & \checkmark & \checkmark & \checkmark & \checkmark & $21.44 \pm 14.89$ & $24.12 \pm 13.75$ \\
\checkmark & \checkmark & \textbf{X} & \checkmark & \checkmark & \checkmark & $17.08 \pm 15.99$ & $19.21 \pm 14.08$ \\
\checkmark & \checkmark & \checkmark & \textbf{X} & \checkmark & \checkmark & $16.89 \pm 15.76$ & $19.57 \pm 14.55$ \\
\checkmark & \checkmark & \checkmark & \checkmark  & \textbf{X} & \checkmark & $ 18.72\pm14.89 $ & $19.86 \pm14.08 $ \\
\checkmark & \checkmark & \checkmark & \checkmark  & \checkmark & \textbf{X} & $ 15.49\pm13.49 $ & $ 19.49\pm14.89 $ \\
\checkmark & \checkmark & \checkmark & \checkmark & \checkmark & \checkmark & \textbf{14.13 $\pm$ 14.69} & \textbf{18.57 $\pm$ 13.37} \\
\bottomrule
\end{tabular}%
}
\end{table*}
The ablation study presented in Table~\ref{tab:ablation} investigates the impact of four key components of the SpectraIrisPAD framework: spectral positional encoding (SPE), token fusion, class-balanced loss, and contrastive loss. These elements are fundamental to the architectural and training novelty proposed in this work, and are directly responsible for enabling generalization to unseen attacks by encoding both spectral-specific and class-discriminative features. Due to page limit constraints and simplicity, we present the ablation study results using training on Artefact~\#1 and testing and remaining artefacts, which includes challenging textured contact lens attacks (Brilliant Blue). 

From the results in Table \ref{tab:ablation}, it is evident that removing any of the four ablated components leads to a measurable performance drop. Token fusion is the most critical, with its removal resulting in the highest D-EER ($21.44\%$), highlighting the role of aggregating cross-spectral token representations for robust decision-making. Excluding SPE leads to a D-EER of $16.92\%$, demonstrating the importance of encoding band-specific positional cues. Omitting the class-balanced loss or contrastive loss impairs the model’s ability to deal with class imbalance and intra-class variation, leading to higher HTER and D-EER values. Similarly, disabling band-adaptive dropout or feature normalisation  increases error, confirming their roles in regularisation and cross-spectral feature stability. The full configuration with all components achieves the best results, confirming their complementary and synergistic contributions to the generalization and robustness of SpectraIrisPAD against diverse PAIs.


\section{Limitations and Future Work}
\label{sec:lim}
This work relies on a purpose-built multispectral iris PAD dataset captured with a specialised sensor, spanning five NIR bands and eight diverse PAIs. While the subject count is smaller than some uni-spectral corpora, this reflects the practical difficulty of collecting large-scale multispectral data with multiple attack instruments under controlled conditions. To our knowledge, there is no public multispectral iris PAD dataset at present, which limits direct cross-dataset benchmarking in the same sensing regime. We partially mitigate this by using a stringent cross-artefact protocol with identity-disjoint splits to probe generalisation across attack types.

We have additionally reported (see Supplementary Material) (i) band-subset analyses using the proposed mask-aware fusion and (ii) zero-shot single-band evaluations to provide insight into transfer when fewer (or only one) wavelengths are available. Nonetheless, two caveats remain. First, single-band testing constitutes a cross-\emph{spectral} transfer scenario for our architecture (which is explicitly designed to exploit inter-band relationships); we therefore avoid any target-domain tuning to keep the setting zero-shot, but this also leaves performance degradation. Second, because quality control (QC) is applied per wavelength, subsets are evaluated on different pools of valid images; while this mirrors deployment, it complicates one-to-one numerical comparisons with the all-band condition.

Future work will address these gaps by: (a) exploring domain adaptation and calibration strategies for single-spectral datasets and unseen sensors, (b) investigating learning-to-weight schemes that are source-trained yet more robust to spectral shifts, (c) studying uncertainty-aware fusion and failure detection under missing/low-quality bands, and (d) developing and, where feasible, releasing resources (code, protocols, and synthetic augmentation tools) to facilitate broader multispectral iris PAD research and reproducibility.

\section{Conclusion}
\label{sec:Conc}
This paper introduced \textbf{SpectraIrisPAD}, a novel and generalizable deep learning framework for multispectral iris Presentation Attack Detection (PAD). By leveraging a DINOv2-based Vision Transformer architecture augmented with spectral positional encoding, token fusion, and contrastive learning, the proposed method effectively captures discriminative features across multiple near-infrared (NIR) spectral bands. To support robust training and benchmarking, we also introduced the \textbf{MSIrPAD} dataset, the most comprehensive multispectral iris PAD dataset to date, comprising over 18,848 samples from eight diverse presentation attack instruments (PAIs), including color-textured lenses, display attacks, and print-based attacks.
Extensive experiments, including cross-artifact evaluations and ablation studies, demonstrate that SpectraIrisPAD consistently outperforms existing state-of-the-art methods in terms of detection accuracy and generalization to unseen attacks. The analysis highlights the importance of spectral diversity and structured token fusion to enhancing PAD performance. The results validate the potential of integrating spectral sensing with transformer-based vision models to build resilient PAD systems for real-world deployment. Future work may explore broader cross-dataset generalization and adaptive spectral attention mechanisms. Future work will explore adapting the proposed framework to cross-sensor settings and integrating additional physiological cues for enhanced security.

{\small
\bibliographystyle{IEEEtran}
	\bibliography{IEEEtran/Iris_PAD_Bib}
}

\end{document}